\documentclass[11pt]{article}
\usepackage[utf8]{inputenc}
\usepackage[T1]{fontenc}
\usepackage{amsmath,amssymb,mathtools,bm,amsthm}
\usepackage[margin=1.25in]{geometry}
\usepackage{hyperref}
\usepackage[capitalise]{cleveref}
\Crefname{assumption}{Assumption}{Assumptions}
\usepackage{color}
\usepackage{nicefrac}
\usepackage{booktabs}
\usepackage{url,graphicx}
\usepackage{algorithmic}
\usepackage{algorithm}
\usepackage{array,multirow}
\usepackage[dvipsnames]{xcolor}
\usepackage{graphicx}
\usepackage{tikz,graphics,color,subcaption}
\usetikzlibrary{positioning,arrows,decorations,decorations.markings,shadows,positioning,arrows.meta,matrix,fit}
\usepackage[sort]{natbib}
\bibpunct[, ]{(}{)}{,}{a}{}{,}%

\newcommand{\norm}[1]{\left\lVert#1\right\rVert}
\newcommand{\pr}{\mathbb{P}}
\newcommand{\expect}{\mathbb{E}}

\newcommand{\ind}{\mathbb{I}}

\newcommand{\bx}{\mathbf{X}}
\newcommand{\bh}{\mathbf{H}}
\newcommand{\bb}{\pmb{\beta}}
\newcommand{\bp}{\pmb{\psi}}

\newcommand{\cf}{\emph{cf.}}
\newcommand{\ie}{\emph{i.e.}}
\newcommand{\eg}{\emph{e.g.}}

\newcommand{\etc}{\emph{etc.}}

\newcommand{\prns}[1]{\left(#1\right)}
\newcommand{\braces}[1]{\left\{#1\right\}}
\newcommand{\bracks}[1]{\left[#1\right]}
\newcommand{\abs}[1]{\left|#1\right|}

\newtheorem{theorem}{Theorem}

\newtheorem{lemma}{Lemma}
\newtheorem{proposition}{Proposition}
\newtheorem{assumption}{Assumption}
\newtheorem{definition}{Definition}

\newenvironment{myproof}[1][Proof]{\begin{proof}[#1]}{\end{proof}}

\title{DTR Bandit:
Learning to Make Response-Adaptive Decisions
With Low Regret}
\author{Yichun Hu,~~~~Nathan Kallus\\Cornell University}
\date{}

\begin{document}

\maketitle

\begin{abstract}
Dynamic treatment regimes (DTRs) are personalized, adaptive, multi-stage treatment plans that adapt treatment decisions both to an individual's initial features and to intermediate outcomes and features at each subsequent stage, which are affected by decisions in prior stages. Examples include personalized first- and second-line treatments of chronic conditions like diabetes, cancer, and depression, which adapt to patient response to first-line treatment, disease progression, and individual characteristics. While existing literature mostly focuses on estimating the optimal DTR from offline data such as from sequentially randomized trials, we study the problem of developing the optimal DTR in an online manner, where the interaction with each individual affect both our cumulative reward and our data collection for future learning. We term this the DTR bandit problem. We propose a novel algorithm that, by carefully balancing exploration and exploitation, is guaranteed to achieve rate-optimal regret when the transition and reward models are linear. We demonstrate our algorithm and its benefits both in synthetic experiments and in a case study of adaptive treatment of major depressive disorder using real-world data.
\end{abstract}

\section{Introduction}

Contextual bandits, where personalized decisions are made sequentially and simultaneously with data collection, are increasingly used to address important decision making problems where data is limited and expensive to collect, with applications in product recommendation \citep{li2011unbiased}, revenue management \citep{qiang2016dynamic}, policymaking \citep{atheytrial}, and personalized medicine \citep{tewari2017ads}.
Bandit algorithms are also increasingly being considered in place of classic randomized trials in order to improve both learning and the outcomes for study participants \citep{villar2015multi,atheytrial}.

But, in many cases decisions are not a one-and-done proposition and there is often a crucial opportunity to adapt to a unit's response to initial treatment.
Generally, a more conservative or more commonly effective treatment is often used at first, reverting to more intensive or more unique treatment if the patient's condition does not respond to the initial intervention or dialing treatment back if the condition is in remission.
\Cref{fig: DTR}, for example, shows a personalized and adaptive treatment plan for the treatment of pediatric attention deficit hyperactivity disorder (ADHD) proposed by \citet{pelham2008evidence}.
These adaptive treatment plans are called \emph{dynamic treatment regimes} \citep[DTRs;][]{murphy2003optimal}.
Crucially, in DTRs, the initial treatment choice not only affects the initial outcome, but also a subject's features in future stages, which may further affect future decisions and outcomes.
Therefore, successful learning of DTRs hinges on accurately learning both the outcome and the transition dynamics and their dependence on subject's features.

\begin{figure}[t!]\centering%
\begin{tikzpicture}[>=latex',font=\footnotesize\linespread{1}\selectfont,
state/.style={draw, shape=rectangle, draw=black, line width=0.5pt, text width=2cm,minimum height=1.25cm,align=center}]
  \node[state,text width=1.25cm](x0){ADHD};
  \node[state](x11)[above right = -0.3cm and 2.75cm of x0]{Low-Dose B-mod};
  \node[state](x12)[below right = -0.3cm and 2.75cm of x0]{Low-Dose Medication};
  \node[state](x21)[above right = -0.3cm and 2.75cm of x11]{Increase B-mod Dose};
  \node[state](x22)[below right = -0.3cm and 2.75cm of x11]{Continue Treatment};
  \node[state](x23)[below right = -0.3cm and 2.75cm of x12]{Augment w/ B-mod};
  \path (x0.east) -- node[sloped] (text1) {Less Severe} (x11.west);
  \path (x0.east) -- node[sloped] (text2) {More Severe} (x12.west);
  \path (x11.east) -- node[sloped] (text3) {Non-response} (x21.west);
  \path (x11.east) -- node[sloped] (text4) {Response} (x22.170);
  \path (x12.east) -- node[sloped] (text5) {Response} (x22.190);
  \path (x12.east) -- node[sloped] (text6) {Non-response} (x23.west);
  \draw[arrows={-Triangle[angle=40:7pt]}]
  (x0.east)--(text1)--(x11.west);
  \draw[arrows={-Triangle[angle=40:7pt]}]
  (x0.east)--(text2)--(x12.west);
  \draw[arrows={-Triangle[angle=40:7pt]}]
  (x11.east)--(text3)--(x21.west);
  \draw[arrows={-Triangle[angle=40:7pt]}]
  (x11.east)--(text4)--(x22.170);
  \draw[arrows={-Triangle[angle=40:7pt]}]
  (x12.east)--(text5)--(x22.190);
  \draw[arrows={-Triangle[angle=40:7pt]}]
  (x12.east)--(text6)--(x23.west);
\end{tikzpicture}
\caption{A simple DTR for the treatment of attention deficit hyperactivity disorder (ADHD). B-mod stands for behavioral modification therapy.}
\label{fig: DTR}
\end{figure}

Traditionally, DTRs are estimated offline, usually from data collected by sequential multiple assignment randomized trials (SMARTs), which randomize assignment at each decision point.
Many \textit{offline} learning methods for DTRs have been proposed, which assume that we have access to the collection of observed trajectories for all patients who participated in the study. 
However, sequential randomized trials are usually costly and limited in size in practice, especially in healthcare domains.
Thus, it would be attractive to use algorithms that learn an effective DTR in an online manner while ensuring subjects have good outcomes, much like the contextual bandit algorithms but being able to be response-adaptive at the unit level. Adaptivity is incredibly crucial in practice: while contextual bandits can personalize treatments to individuals' baseline characteristics such as age, weight, comorbidities, \etc, these features' predictive ability pales in comparison to the informativeness of an individual's actual response to an actual treatment.

In this paper, we study efficient online learning algorithms for optimal DTRs when the $Q$ functions are linear.
As in contextual bandits, from each unit we interact with, we only observe \textit{bandit feedback}, \ie, only the rewards of the actions taken and never that of the others, so we face the trade-off between exploration and exploitation:
we are motivated to apply the current-best DTR so to collect the highest reward for the current unit, but we also need to explore other treatments for fear of missing better options for future units due to lack of data.
Unlike contextual bandits, our decisions not only impact the immediate reward and which arm we observe at the present context for use in future learning, but they \emph{also} impact both the possible reward in future stages for the same unit \emph{and} the possible context at which we may obtain new data on future-stage reward structure.
Our proposed algorithm tackles the trade-off between exploration and exploitation by carefully balancing the two imperatives and maintaining two sets of parameter estimators (one unbiased but imprecise, one biased but much more precise) to use in different context regions \citep[inspired by][]{goldenshluger2013linear}.
We prove that, under a sharp margin condition (\ie, $\alpha = 1$ in \cref{assumption: margin} below), our algorithm only incurs $O(\log T)$ regret after interacting with $T$ units, compared to the oracle policy that knows the true reward and transition models exactly.
Our result is rate optimal and comparable to regret rates
in the one-stage linear-response bandit problem where we may treat each unit only once.

\subsection{Related Literature} \label{section: literature}

\paragraph{Dynamic Treatment Regimes.}
DTRs generalize individualized treatment regimes \citep[ITRs;][]{kallus2018balanced,athey2021policy,qian2011performance,zhao2012estimating}, which personalize single treatment interventions based on baseline unit features.
In DTR models, additional interventions are made and can be tailored to the patient's time-varying features, which may be impacted by previous interventions.
A number of methods have been proposed to estimate the optimal DTR from batched data from either randomized trials or observational studies, including potential outcome imputation \citep{murphy2001marginal}, likelihood-based methods \citep{thall2007bayesian}, $A$-learning \citep{murphy2003optimal,schulte2014q}, and weight-based methods \citep{zhao2015new,zhang2013robust,jiang2019entropy}, among others.
Our work leverages ideas from $Q$-learning, which has been used for learning DTRs offline from both randomized trial data \citep{zhao2009reinforcement,chakraborty2010inference,shortreed2011informing,song2015penalized} and observational data \citep{moodie2014q}.
For a more complete review of DTR methods, see \citet{chakraborty2013statistical}.

\paragraph{Efficient Reinforcement Learning (RL).}
Online algorithms and regret analysis appear rarely in the DTR literature.
A closely related topic is studied in regret-efficient exploration in episodic RL problems.
The tabular RL case, where both state and action spaces are finite, is well-studied in the literature \citep[\eg,][]{azar2017minimax}.
However, the state space (which corresponds to units' features in DTR problems) is usually enormous or even infinite in real-world applications, which makes the tabular algorithms (whose regret grows polynomially in the number of states) untenable in practice.
Thus, effectively modeling the problem structure is necessary.
{\citet{jin2019provably} proposed an efficient RL algorithm with linear function approximation (LSVI-UCB) and proved $O(\sqrt{T})$ regret.
The current work differs from \citet{jin2019provably} by introducing the margin condition to the DTR problem, which allows for much lower regret rate when feature distributions are not arbitrarily concentrated near the decision boundary, as would indeed not occur in practice.
In the context of RL with function approximation, our paper is the \emph{first} to leverage such a condition to obtain logarithmic regret rates, which are faster than square-root rates that are minimax optimal in the absence of a margin condition.
Moreover, we consider the \textit{new} problem of delayed feedback in \cref{section: delay}.
Beyond regret minimization, some recent work studied other aspects of RL with linear/low-rank MDPs, including policy learning \citep{yang2019sample} and representation learning \citep{agarwal2020flambe}.}

\paragraph{Contextual Bandits.}
Contextual bandits extends multi-armed bandits \citep{lai1985asymptotically} by introducing observations of side information at each time point \citep[\eg,][]{auer2002using}.
While some literature makes no assumption about the context generating process and allows for adversarially chosen contexts \citep[\eg,][]{beygelzimer2011contextual}, this can be overly pessimistic in real-world scenarios and inevitably leads to high regret. 
Thus, another line of literature considers the stochastic contextual bandit problem, which assumes that contexts and rewards are drawn i.i.d. from a stationary but unknown distribution \citep[\eg,][]{dudik2011efficient}.
We will focus on the stochastic setting in this paper.
The margin condition, which has been introduced to characterize the hardness of classification problems \citep{mammen1999smooth}, has been used in stochastic contextual bandits in \citet{rigollet2010nonparametric,goldenshluger2013linear,hu2022smooth,bastani2020online}, among others.
The margin condition parameter (usually denoted as $\alpha\geq0$) quantifies the concentration of contexts near the decision boundary between arms and crucially determines the learning difficulty of bandit problems.
Among these, \citet{rigollet2010nonparametric,perchet2013multi,hu2022smooth} assume non-parametric relationships between rewards and covariates.
Nonparametric algorithms may perform poorly in moderate to high dimensions due to the curse of dimension, so much of the literature focuses on the case where the rewards are parametric -- particularly, linear -- functions of the observed contexts.
Under the linear model, \citet{goldenshluger2013linear} achieve optimal $O(\log T)$ regret under a sharp margin ($\alpha = 1$).
\citet{bastani2021mostly} show that the analysis techniques can be easily extended to general margin conditions.

\subsection{Organization}
The rest of the paper is organized as follows. In  \cref{section: formulation}, we formally introduce the DTR bandit problem and assumptions. We describes our proposed algorithm in \cref{section: algorithm}. In \cref{section: theory}, we analyze our algorithm theoretically: we derive an upper bound on the regret of our algorithm, and we show a matching lower bound on the regret of any algorithm. In \cref{section: empirics} we evaluate the empirical performance of our algorithm on both synthetic data and a real-world sequenced randomized trial on treatment regimes for patients with major depressive disorder. We conclude our paper and discuss future directions in \cref{section: conclusion}. 

\subsection{Notation}
We use $\|\cdot\|$ to represent the Euclidean norm of vectors and the spectral norm of matrices, $\|\cdot\|_1$ to represent the $L_1$ norm of vectors, and $\|\cdot\|_{\infty}$ to represent the infinity norm of vectors.
For any integer $n$, let $[n]$ denote the set $\{1,\dots, n\}$.
For two scalers $a, b \in \mathbb{R}$, let $a \wedge b = \min\{a, b\}$ and $a \vee b = \max\{a, b\}$.
For an event $A$, the indicator variable $\ind(A)$ is equal to $1$ if $A$ is true and $0$ otherwise.
For a set $S$, its cardinality is denoted as $|S|$.
For a matrix $\mathcal{A}$, its minimum and maximum eigenvalues are denoted as $\lambda_{\min}(\mathcal{A})$, $\lambda_{\max}(\mathcal{A})$, respectively.
For two functions $f_1(T) > 0$ and $f_2(T) > 0$, we use the standard notation for asymptotic order: $f_1(T) = O(f_2(T))$ represents $\limsup_{T \to \infty} f_1(T)/f_2(T) < \infty$, $f_1(T) = \Omega(f_2(T))$ represents $\liminf_{T \to \infty}  f_1(T)/f_2(T) > 0$, and $f_1(T) = \Theta(f_2(T))$ represents simultaneously $f_1(T) = \Omega(f_2(T))$ and $f_1(T) = O(f_2(T))$.

\section{The DTR Bandit Problem} \label{section: formulation}
We now formally formulate the DTR bandit problem. For simplicity,  we focus on the two-armed two-stage DTR bandits. We present the extension to the general multi-armed problem in \cref{section: k arms} and discuss the $M$-stage problem in \cref{section: conclusion}.

\subsection{Problem Setup} \label{section: problem setup}

The DTR bandit problem proceeds in rounds indexed by $t$.
{
For $t = 1, 2, \dots$,
nature draws a potential trajectory $\braces{\bx_{t,1}, \bx_{t,2}\prns{a_1}, \bx_{t,3}\prns{a_1, a_2}}_{a_1\in[2], a_2\in[2]}$ i.i.d. from a fixed but unknown distribution $\mathcal{P}$. 
In the first stage of each round, the decision maker observes the initial context $\bx_{t,1}\in \mathcal{X}_1\subseteq \mathbb{R}^{p_1}$ and pulls an arm $A_{t,1}\in [2]$ according to the observed context and the data collected so far. 
They then obtain an updated context $\bx_{t,2} = \bx_{t,2}(A_{t,1})\in \mathcal{X}_2\subseteq \mathbb{R}^{p_2}$.
In the second stage, the decision maker pulls an arm $A_{t,2}\in [2]$ according to the updated information set so far, and receives the final context $\bx_{t,3} = \bx_{t,3}(A_{t,1}, A_{t,2}) \in\mathcal{X}_3 \subseteq \mathbb{R}^{p_3}$.
The reward the decision maker gets from time $t$ is 
$$Y_t = f\prns{\bx_{t,1}, A_{t,1}, \bx_{t,2}, A_{t,2}, \bx_{t,3}},$$
where $f$ is a known function that aggregates all contexts and actions in this round into a scalar.
To emphasize the dependence on actions, we write $Y_t(a_1, a_2)$ as the potential reward had they taken $a_1$ in stage one and $a_2$ in stage two, \ie,
\begin{align*}
    Y_t(a_1, a_2) = f\prns{\bx_{t,1}, a_1, \bx_{t,2}\prns{a_1}, a_2, \bx_{t,3}\prns{a_1, a_2}},
\end{align*}
so $Y_t = Y_t(A_{t,1}, A_{t,2})$.
Besides, we define the history prior to each stage as $\bh_{t,1} = \bx_{t,1}$ and $\bh_{t,2} = \prns{\bx_{t,1}^\intercal, A_{t,1}, \bx_{t,2}^\intercal}^\intercal$, and the potential history prior to stage two had we pulled arm $a_1$ in stage one as $\bh_{t,2}(a_1) = \prns{\bx_{t,1}^\intercal, a_1, \bx_{t,2}^\intercal(a_1)}^\intercal$; we let $\mathcal{H}_1 = \mathcal{X}_1$ and the support of $\bh_{t,2}$ be $\mathcal{H}_2$.
}

Specifically, the decision maker aims to maximize cumulative rewards at any horizon $T$, namely $\sum_{t=1}^T Y_t$, using an admissible policy (interchangeably called algorithm or allocation rule). An admissible policy, $\mathcal{A} = \{\mathcal{A}_{t,1}, \mathcal{A}_{t,2}\}$, is a sequence of \emph{random} functions $\mathcal{A}_{t,1}: \mathcal{H}_1 \to [2], \mathcal{A}_{t,2}: \mathcal{H}_2 \to [2]$
such that, for each $t$, $\mathcal{A}_{t,1}$ is conditionally independent of $\{\bx_{t',1},A_{t',1},\bx_{t',2}(a_1), A_{t',2}, \bx_{t',3}(a_1, a_2)\}_{t'\in [T], a_1\in [2], a_2\in [2]}$ given $\{\bx_{t',1},A_{t',1},\bx_{t',2}, A_{t',2}, \bx_{t',3}\}_{t'\in [t-1]}$, and $\mathcal{A}_{t,2}$ is conditionally independent of $\{\bx_{t',1},A_{t',1},\bx_{t',2}(a_1), A_{t',2}, \bx_{t',3}(a_1, a_2)\}_{t'\in [T], a_1\in [2], a_2\in [2]}$ given $\{\bx_{t',1},A_{t',1},\bx_{t',2}, A_{t',2}, \bx_{t',3}\}_{t'\in [t-1]} \cup \{\bx_{t,1},A_{t,1},\bx_{t,2}\}$, where we let $A_{t',1}=\mathcal{A}_{t',1}(\bx_{t',1})$, $\bx_{t',2}=\bx_{t',2}(A_{t',1})$, $A_{t',2}=\mathcal{A}_{t',2}(\bh_{t',2})$, and $\bx_{t',3}=\bx_{t',3}(A_{t',1},A_{t',2})$.

\subsection{$Q$-Functions and Regret}
We can use $Q$-functions \citep[\cf][]{sutton2018reinforcement} to conveniently represent the expected total rewards in future stages of each context-action pair if one were to follow the optimal policy. Specifically,
we let
{
\begin{align} 
\label{eq: q2}
 Q_2(\mathbf{h}_2,a_2) &= \expect\bracks{Y_t\mid \bh_{t,2} = \mathbf{h}_2, A_{t,2} = a_2},
 \\
 \label{eq: q1}
  Q_1(\mathbf{x}_1,a_1)
   &= \expect\bracks{\max_{a_2\in[2]} Q_2\prns{\bh_{t,2},a_2}\mid \bx_{t,1} = \mathbf{x}_1, A_{t,1} = a_1}.
\end{align}
}
Then $Q_2(\mathbf{h}_2,a_2)$ represents the expected reward if we pull arm $a_2$ in stage 2 given history $\mathbf{h}_2$, while $Q_1(\mathbf{x}_1,a_1)$ represents the expected reward if we pull arm $a_1$ in stage 1 given initial context $\mathbf{x}_1$ and then, when the resulting stage 2 context realizes, pull the arm in stage 2 that maximizes expected reward.

Had the decision maker known the true functions $Q_1$ and $Q_2$ (but not the realizations of {$\bx_{t,2}(a_1), \bx_{t,3}(a_1, a_2)$} before they are observed), the optimal decision at time $t$ would be to follow the oracle policy $\mathcal A^*$ that always pulls the arm with higher $Q$-function in each stage, \ie,
$${A_{t,1}^*}\triangleq\mathcal A^*_{t,1}(\bx_{t,1}) = \arg \max_{a_1\in [2]} Q_1(\bx_{t,1},a_1),$$
$${A_{t,2}^*}\triangleq\mathcal A^*_{t,2}(\bh_{t,2}^*) =  \arg \max_{a_2\in [2]} Q_2(\bh_{t,2}^* ,a_2),$$
where {$\bh_{t,2}^* = \prns{\bx_{t,1}^\intercal, A^*_{t,1}, \bx_{t,2}^\intercal(A_{t,1}^*)}^\intercal$}.
(Notice $A_{t,1}^*,A_{t,2}^*$ are random variables.)
The average reward they would then get at each round (and, in particular, in round $t$) would be
{
\begin{align*}
    \expect[Y_t\prns{A_{t,1}^*, A_{t,2}^*}].
\end{align*}
}
However, because they do not know these $Q$-functions, this optimal reward is infeasible in practice. The expected reward they would get in round $t$ by following algorithm $\mathcal{A}$ is
{
\begin{align*}
    \expect[ Y_t(A_{t,1}, A_{t,2})],
\end{align*}
}
where $A_{t,1}=\mathcal{A}_{t,1}(\bx_{t,1})$, $\bx_{t,2} = \bx_{t,2}\prns{A_{t,1}}$, and $A_{t,2}=\mathcal{A}_{t,2}(\bh_{t,2})$. $\bx_{t,2}$ may differ from $\bx_{t,2}\prns{A_{t,2}^*}$, because the decision maker may pull the suboptimal arm in the first stage.

Define the per-step regret $r_t$ as the difference between these two average rewards:
{
$$r_t \triangleq \expect[Y_t\prns{A_{t,1}^*, A_{t,2}^*}] - \expect[ Y_t(A_{t,1}, A_{t,2})].$$
}
We measure the performance of an algorithm $\mathcal{A}$ by its \textit{expected cumulative regret} compared to the oracle policy $\mathcal A^*$ up to time $T$, which quantifies how much $\mathcal{A}$ is inferior to $\mathcal A^*$:
$$R_T(\mathcal{A}) = \sum_{t=1}^T r_t.$$
The growth of this function in $T$ quantifies the quality of $\mathcal{A}$.

While the expected cumulative regret can be seen as the difference between two marginal expectations in two independent stochastic systems, it is often helpful to couple them via the context realizations in order to analyze the regret. 
For example, in contextual bandits, one can rephrase the per-step expected regret as the expected mean-reward difference at each context. 
It turns out $Q$-functions are helpful in this respect as well.
The following proposition shows that $r_t$ can actually be equivalently written as the sum of the expected differences in each of the $Q$-functions.
{

\begin{proposition}\label{prop: expected regret}
For any $t\in [T]$,
$$r_t = \expect\bracks{Q_1(\bx_{t,1}, A^*_{t,1})-Q_1(\bx_{t,1}, A_{t,1}) + \max_{a_2}Q_2(\bh_{t,2}, a_2)- Q_2(\bh_{t,2}, A_{t,2})}.$$
\end{proposition}
}
Intuitively, \cref{prop: expected regret} shows that the per-step expected regret comes from two sources: the total regret over both stages that we expect to get due to a suboptimal decision in the first stage (even if we follow the optimal policy afterwards) and the regret we expect to get due to a suboptimal decision in the second stage.

\subsection{Assumptions}
We now introduce several assumptions that rigorously define our problem instances. 
We first introduce the notion of sub-Gaussianity, which will be used repeatedly.
\begin{definition}[Sub-Gaussian random variables] \label{def: subgaussian}
A random variable $Z\in \mathbb{R}$ is $\sigma$-sub-Gaussian if $\expect[e^{s Z}]\le e^{\sigma^2 s^2/2}$ for $\forall s\in \mathbb{R}$.
\end{definition}
This definition implies $\expect Z = 0$ and $Var[Z] \le \sigma^2$. Any bounded $0$-mean random variable $Z$ with $|Z|\le z_{\max}$ is $z_{\max}$-sub-Gaussian. Another classic example of a $\sigma$-sub-Gaussian random variable is the centered Gaussian distribution with variance $\sigma^2$.
{

\paragraph{Linear $Q$-Functions}
Our first assumption states that the $Q$-functions are linear in some feature maps of the history information, which is the typical modeling choice for $Q$-functions in the DTR literature \citep{chakraborty2013inference, chakraborty2013statistical}. 
\begin{assumption}\label{assumption: bounded}
For $m\in[2], a\in[2]$, there exists some known function $\bp_m: \mathcal{H}_m \rightarrow \mathbb{R}^{d_m}$ and some unknown parameter $\bb_{a,m}$ such that
\begin{align*}
    Q_m(\mathbf{h}_m, a) = \bb_{a,m}^\intercal \bp_m(\mathbf{h}_m).
\end{align*}
Moreover, there exist positive constants $x_{\max}, b_{\max}, \sigma$ such that $\norm{\bp_m(\mathbf{h}_m)}\le x_{\max}$ for all $\mathbf{h}_m\in \mathcal{H}_m$, $\norm{\bb_{a,m}}\le b_{\max}$, and the residues
\begin{align*}
    \eta^{(a_1)}_{t,1} & = \max_{a_2\in[2]} Q_2\prns{\bh_{t,2}(a_1), a_2} - Q_1(\bx_{t,1}, a_1), \\
    \eta^{(a_1, a_2)}_{t,2} & =Y_t(a_1, a_2) -Q_2(\bx_{t,2}(a_1), a_2)
\end{align*}
are $\sigma$-sub-Gaussian and independent of all else. 
\end{assumption}
}

Boundedness of ferature maps and parameters is necessary to guarantee that the per-step regret is finite in worst-case over instances, which ensures that exploration will not be uncontrollably costly. This assumption is common in literature \citep{goldenshluger2013linear, bastani2020online} and is usually satisfied, because most attributes in practice are bounded in nature.

\paragraph{Margin Condition.}
Our second assumption is an adapted version of the margin condition commonly used in stochastic contextual bandits \citep{rigollet2010nonparametric,goldenshluger2013linear} and classification \citep{mammen1999smooth}. It determines how the estimation error of $Q$-functions translates into regret of decision-making.
\begin{assumption}\label{assumption: margin}
There exist positive constants $\gamma_1, \gamma_2, \alpha_1, \alpha_2$ such that for any $\kappa>0$,
\begin{align} \label{eq: margin 1}
   &\mathbb{P}\prns{0<\abs{Q_1(\bx_{t,1},1)-Q_1(\bx_{t,1},2)} \le \kappa}\le \gamma_1 \kappa^{\alpha_1},
\\
\label{eq: margin 2}
    &\mathbb{P}\prns{0<\abs{Q_2\prns{{\bh_{t,2}(a_1)}, 1} - Q_2\prns{{\bh_{t,2}(a_1)}, 2}} \le \kappa} \le \gamma_2 \kappa^{\alpha_2}\quad\forall a_1\in [2].
\end{align}
To simplify notation we let $\alpha = \min\{\alpha_1, \alpha_2\}$.
\end{assumption}

Differently from existing literature, we impose the margin condition on the difference between $Q$-functions, rather than the mean reward functions.
Moreover, because the second-stage covariate distribution is directly affected by the implemented algorithm, we explicitly require the margin condition to hold in the second stage for each initial-stage arm.

Intuitively, the margin condition quantifies the concentration of contexts very near the decision boundary, which measures the difficulty of determining the optimal arm.
When either $\alpha_1$ or $\alpha_2$ is very small, the $Q$-functions in the corresponding stage can be arbitrarily close to each other with high probability, so even very small estimation error may lead to suboptimal decisions.
In contrast, when both $\alpha_1$ and $\alpha_2$ are very large, either the $Q$-functions are the same so that pulling which arm does not matter, or they are separated apart from each other with sufficient probability so that the optimal arm is easy to identify.

{

The next lemma shows that, for many continuous distributions with upper bounded density, we have $\alpha = 1$.
In fact, \cite{goldenshluger2013linear,bastani2020online} explicitly assume $\alpha = 1$.
\begin{lemma} \label{lemma: alpha is one}
Suppose each of $\bp_1(\bx_{t,1})$, $\bp_2(\bh_{t,2}(1))$ and $\bp_2(\bh_{t,2}(2))$ has a density and this density is bounded by $\mu_{\max}$.
Then, \cref{assumption: margin} holds with $\alpha_m = 1$ and $\gamma_m = 12\mu_{\max} x_{\max}^d / \norm{\bb_{1,m} - \bb_{2,m}}$.

\end{lemma}
}

\paragraph{Positive-definiteness of Design Matrices.}
Our last assumption is the positive-definiteness of design matrices in both stages, and we focus on decision regions where arm $a$ is optimal by a margin. Our \cref{assumption: diversity} is closely related to asumption $(A3)$ in \citet{goldenshluger2013linear} and assumptions $3$ and $EC.1$ in \citet{bastani2020online}.

{

\begin{assumption}\label{assumption: diversity}
There exist positive constants $\Delta_1, \Delta_2, \phi_1, \phi_2$ such that the sets of histories where each arm is optimal in each stage, 
$$U_{a,m} \triangleq \{\mathbf{h}\in \mathcal{H}_m: Q_m(\mathbf{h},a) > Q_m(\mathbf{h},3-a)+\Delta_m\}, \quad a,m \in [2],$$
satisfy for all $a\in [2]$:
\begin{align}
  &\lambda_{\min}\prns{\expect\bracks{ \bp_1\prns{\bx_{t,1}}\bp_1^\intercal\prns{\bx_{t,1}}\ind\braces{ \bx_{t,1}\in U_{a_1,1}}}}\ge \phi_1, \label{assumption3: 1}\\
  & \lambda_{\min}\prns{{\textstyle\sum_{a_1\in[2]}}\expect\bracks{ \bp_2\prns{\bh_{t,2}(a_1)}\bp_2^\intercal\prns{\bh_{t,2}(a_1)}\ind\braces{ \bh_{t,2}(a_1) \in U_{a,2}, \bx_{t,1}\in U_{a_1,1}}}}\ge \phi_2. \label{assumption3: 2}
\end{align}
\end{assumption}
}
In the first stage, we require that the design matrices of $\bx_{t,1}$'s \textit{in the optimal regions} are positive-definite (\cref{assumption3: 1}).
In the second stage, we require that 
the design matrices of $\bh_{t,2}(a_1)$'s \textit{in the optimal regions} in the second stage which come from pulling the optimal arm \textit{in the optimal regions} in the first stage are also positive-definite (\cref{assumption3: 2}).
These are common technical requirements for the uniqueness and convergence of OLS (ordinary least squares) estimators.
Intuitively, \cref{assumption: diversity} implies that, as long as we pull arm $a$ in stage $m$ when the context falls in $U_{a,m}$, it is possible to collect sufficient informative samples (of order $\Theta(t)$ at time $t$) on both arms without necessarily forcing exploration too often, and in turn we can quickly improve estimation accuracy without collecting too much regret.
And, as long as we have moderately accurate estimation of the parameters, we are very likely to pull the right arm in the $U_{a,m}$ regions.
\cref{assumption3: 1,assumption3: 2} also imply that for each arm there is positive probability for it to be optimal by a margin: 
\begin{lemma} \label{lemma: probability bound}
When \cref{assumption: diversity} holds, there exists $p, p'>0$ such that for all $a\in [2]$:
\begin{align}
  & \mathbb{P}\prns{\bx_{t,1}\in U_{a,1}} \ge p, \label{lemma1: 1}\\
  & {\textstyle\sum_{a_1\in[2]}}\mathbb{P}\prns{\bh_{t,2}(a_1) \in U_{a,2}, \bx_{t,1}\in U_{a_1,1}} \ge p'. \label{lemma1: 2}
\end{align}
\end{lemma}
In real-world applications, we indeed often expect each of the treatments to be optimal at least in some cases.
Nonetheless, we will discuss a relaxation of \cref{assumption: diversity} to allow for arms that are suboptimal everywhere in \cref{section: k arms}.
Our algorithm for two arms is unchanged; we only extend the analysis.

\section{Algorithm} \label{section: algorithm}
In this section we present our algorithm for the DTR bandits.
Our algorithm adapts an idea that originated in the one-stage problem \citep{goldenshluger2013linear} to the sequential setting: specifically, we maintain \emph{two} sets of $Q$-function estimators at each time point, one based solely on samples from \textit{forced (randomized) pulls} at specially prescribed time steps and the other based on \textit{all samples} up to the current round, which includes rounds in which we pull the arms that appear to be optimal based on past data, inducing a dependence structure in the data. Which of the two estimates we use when determining which arm appears better depends on how close we appear to be to the decision boundary. We discuss the intuition and importance of this below.

\subsection{Forced-Pull Estimators and All-Samples Estimators}

Our algorithm will occasionally force pull certain arms without regard to contextual observations.
Fix $q\in \mathbb{Z}^+$ as our forced sampling schedule parameter.
Similar to \citet{bastani2020online}, we prescribe the following time steps to perform force pulls: { for $a_1, a_2\in [2]$,
\begin{equation}\label{eq: forced samples}
 \mathcal{T}_{(a_1, a_2)} \triangleq \{(2^{n+2}-4)q+j~:~j=q(2a_1 + a_2-3)+1,q(2a_1 + a_2-3)+2, \dots, q(2a_1 + a_2-2),\ n=0,1,2,\dots\}.
\end{equation}
Then, at time step $t$, if $t\in \mathcal{T}_{(a_1, a_2)}$, we pull $a_1$ in stage one and $a_2$ in stage two regardless of what history we see. Moreover, we denote by $\mathcal{T}_{a,1}(t) \triangleq \prns{\cup_{a_2\in[2]} \mathcal{T}_{(a, a_2)}}  \cap [t]$ and $\mathcal{T}_{a,2}(t) \triangleq \prns{\cup_{a_1\in[2]} \mathcal{T}_{(a_1, a)}}  \cap [t]$ the time indices where we force pull arm $a$ in each stage, up to time $t$. As we will show in \cref{lemma: forced pull numbers}, $|\mathcal{T}_{a,m}(t)|$ is of order $\Theta (q\log t)$.}

Based solely on forced samples, we get the forced-sample estimators. 
{
Define
\begin{align*}
    \tilde{\Sigma}_{a,m}(t) &= \sum_{j\in \mathcal{T}_{a,m}(t)}\bp_m\prns{\bh_{j,m}}\bp_m^\intercal\prns{\bh_{j,m}} \quad a,m\in[2],\\
    \tilde{\bb}_{a, 2}(t) &=\tilde{\Sigma}_{a,2}^{-1}(t) \prns{\sum_{j\in \mathcal{T}_{a,2}(t)}\bp_2(\bh_{j,2})Y_j } \quad a\in[2].
\end{align*}
Using $\tilde{\bb}_{a, 2}(t)$, we can impute pseudo-outcomes
\begin{align*}
    \tilde{Y}_j = \max_{a\in[2]}~ \tilde{\bb}_{a, 2}^\intercal(t) \bp_2\prns{\bh_{j,2}} \quad j\in [t],
\end{align*}
and estimate the first-stage parameter
\begin{align*}
    \tilde{\bb}_{a, 1}(t) = \tilde{\Sigma}_{a,1}^{-1}(t) \prns{\sum_{j\in \mathcal{T}_{a,1}(t)}\bp_1(\bx_{j,1})\tilde{Y}_j } \quad a\in[2].
\end{align*}
Finally, we can compute the following forced-pull $Q$ estimators:}
\begin{align*}
    {\tilde{Q}_{t,m} (\mathbf{h}, a) =  \Tilde{\bb}_{a,m}^\intercal(t) \bp_m(\mathbf{h}).}
\end{align*}

Next we discuss the $Q$-estimators constructed from \emph{all} past samples.
For $a,m\in[2]$, define $\mathcal{S}_{a,m}(t)\triangleq\{j\in [t]: A_{j,m} = a\}$ to be all rounds in which we pull arm $a$ in the $m$-th stage, up to time $t$. In contrast to the forced-pull estimators, the all-samples estimators are based on all samples collected:
{
\begin{align*}
    \hat{\Sigma}_{a,m}(t) & =  \sum_{j\in \mathcal{S}_{a,m}(t)}\bp_m\prns{\bh_{j,m}}\bp_m^\intercal\prns{\bh_{j,m}} \quad a,m\in[2],\\
    \hat{\bb}_{a, 2}(t) & =\hat{\Sigma}_{a,2}^{-1}(t) \prns{\sum_{j\in \mathcal{S}_{a,2}(t)}\bp_2(\bh_{j,2})Y_j } \quad a\in[2],\\
     \hat{Y}_j & = \max_{a\in[2]} \hat{\bb}_{a, 2}^\intercal(t) \bp_2\prns{\bh_{j,2}} \quad j\in [t],\\
     \hat{\bb}_{a, 1}(t) & = \hat{\Sigma}_{a,1}^{-1}(t) \prns{\sum_{j\in \mathcal{S}_{a,1}(t)} \bp_1(\bx_{j,1})\hat{Y}_j  } \quad a\in[2].
\end{align*}
Using these, we can compute the all-sample $Q$ estimators:}
\begin{align*}
    {\hat{Q}_{t,m} (\mathbf{h}, a) =  \hat{\bb}_{a,m}^\intercal(t) \bp_m(\mathbf{h}).}
\end{align*}

\subsection{Running the Algorithm} \label{section: running alg}

We now describe how the algorithm proceeds. The algorithm has three parameters: the scheduling parameter $q$ and the covariate diversity parameters $\Delta_1,\Delta_2$.
(We discuss their choice in our empirical analyses in \cref{section: empirics}.)
First,
at time $t$, if $t$ falls into the forced sampling steps $\mathcal{T}_{(a_1, a_2)}$, we pull $a_1$ in stage one and $a_2$ in stage two regardless of $\bh_{t,1}, \bh_{t,2}$.
Otherwise, if $t\notin \cup_{a_1, a_2\in[2]}\mathcal{T}_{(a_1, a_2)}$, we use our $Q$-estimators to choose the action in each stage, using the forced-pull estimators if they provide a clear enough distinction between arms and otherwise using the all-samples estimators.
Specifically, in the $m$-th stage, we first compare the $\Tilde{Q}_{t-1,m}$ estimators of both arms, and if they differ by at least $\Delta_m/2$ at the observed history prior to the stage, $\bh_{t,m}$, then we pull the arm with higher $\Tilde{Q}_{t-1,m}$ value.
Otherwise, if the $\Tilde{Q}_{t-1,m}$ values are too close, we pull the arm with higher $\hat{Q}_{t-1,m}$ value at $\bh_{t,m}$.
At the end of time $t$, we update all estimators and enter the next round.
We summarize our algorithm in \cref{alg}.

\begin{algorithm}[t!]%
    \caption{\textsc{DTRBandit}}
    \label{alg}
    \begin{algorithmic}[1]
    \item \textbf{Input parameters:} $\Delta_1, \Delta_2, q$.
    \FOR{$t = 1,2,\dots$}
                \IF{{$t \in \mathcal{T}_{(a_1, a_2)}$ for $a_1, a_2 \in[2]$}}
                \STATE{{pull $A_{t,1} = a_1, A_{t,2} = a_2$} }
                \STATE{observe rewards and compute $\hat{\bb}_{a,m}(t)$, $\Tilde{\bb}_{a,m}(t)$ for $a,m\in[2]$}
                \ELSE
                \FOR{$m = 1,2$}
                \IF{$\abs{\Tilde{Q}_{t-1,m}(\bh_{t,m},1)-\Tilde{Q}_{t-1,m}(\bh_{t,m},2)}>\Delta_m/2$}
                \STATE{pull $A_{t,m} = \arg\max_{a=1,2} \Tilde{Q}_{t-1,m}(\bh_{t,m},a)$}
                \ELSE
                \STATE{pull $A_{t,m} = \arg\max_{a=1,2} \hat{Q}_{t-1,m}(\bh_{t,m},a)$}
                \ENDIF
                \STATE{observe rewards and compute $\hat{\bb}_{a,m}(t)$ for $a\in[2]$}
                \ENDFOR\\
                \ENDIF
    \ENDFOR
    \end{algorithmic}
\end{algorithm}

\subsection{Explaining the Algorithm} \label{section: explain}
We next give some intuition for our choices and for why to expect that \cref{alg} should have low regret.
Forced pulls offer clean, independent samples on both arms, and the second-stage OLS estimator based on them is unbiased and known to concentrate around the true parameter values.
{
This also ensures that our $\tilde{Y}$ estimate is close to $ \max_{a_2\in[2]} Q_2\prns{\bh_{2}(a_1), a_2}$, which leads to the concentration of the first-stage OLS estimator.}

All these factors result in the concentration of the forced-pull $Q$-estimators around true $Q$-functions, and as we show in \cref{prop: convergence of Tilde Q}, we only need $\Theta(\log t)$ independent samples (as our forced-pull schedule ensures) to get a uniform $\Delta_m/4$-approximation of $Q_m$ with probability $O\prns{t^{-2}}$, for $m=1,2$.
For each action, however, the $Q$-functions are not perfectly separated and they may be arbitrarily close, making it hard to choose the right action unless our estimates are well separated.
When our approximation error on $Q_m$ is at most $\Delta_m/4$, our error on the difference $Q_m(\mathbf{h},1)-Q_m(\mathbf{h},2)$ is at most $\Delta_m/2$. Thus, whenever $\abs{\Tilde Q_m(\mathbf{h},1)-\Tilde Q_m(\mathbf{h},2)}>\Delta_m/2$, we are (almost) certain about which arm is optimal.

Moreover, as long as our $\Delta_m/4$-approximation applies,
whenever $\abs{Q_m(\mathbf{h},1)-Q_m(\mathbf{h},2)}>\Delta_m$, \ie, $\mathbf{h}\in U_{1,m}\cup U_{2,m}$, we also expect to fall into such a scenario where the forced-samples estimators are $\Delta_m/2$-separated. Under \cref{assumption: diversity}, this occurs with positive probability.
Therefore, following our algorithm in these regions, we expect to collect $\Theta(t)$ samples on each arm in each stage, but only for contexts far from the margin. \Cref{assumption: diversity} nonetheless ensures sufficient diversity in that region.
Since this offers \emph{much} more data than the very limited $\Theta(\log t)$ forced pulls,
we get a much faster (of rate $\sqrt{t}$) concentration for the all-sample estimator, $\hat{Q}$. This allows us to make very accurate decisions near the margin, where the forced-pull estimator is not accurate enough.

\section{Theoretical Guarantees} \label{section: theory}
In this section, we provide a regret upper bound for \cref{alg} as well as a matching (in order of $T$) lower bound on the regret of any other algorithm, which shows that our algorithm is rate optimal.
We provide an outline of the proof of the below in \cref{sec:upperboundanalysis}, where we break up the argument into modular Propositions, each of which we prove in detail in \cref{section: proof}.

\begin{theorem}\label{thm: regret upper bound}
Let $\hat{\mathcal A}$ denote our algorithm described in \cref{alg}. Suppose \cref{assumption: bounded,assumption: margin,assumption: diversity} hold.
Then there exists a constant $\tilde{C}$ such that, if $q\ge \tilde{C}$, then the expected regret of our
algorithm is bounded as follows:
\begin{equation}R_T(\hat{\mathcal{A}}) =
\begin{cases}
O( d^{\frac{\alpha+1}{2}}(\log d)^{\frac{\alpha+2}{2}} T^{\frac{1-\alpha}{2}} +  d \log d \log T + (d\log d)^2),& \alpha <1\\
O(d (\log d)^{\frac{3}{2}} \log T + (d\log d)^2),& \alpha =1 \\
O(d \log d \log T + (d\log d)^2 +  d^{\frac{\alpha+1}{2}}(\log d)^{\frac{\alpha+2}{2}}),& \alpha >1.
\end{cases}\end{equation}
More specifically, letting \begin{align*}
    & \Tilde{C}_1 =\frac{16 (\log d+2) x_{\max}^2\prns{ \prns{128 d x_{\infty}^2 \sigma^2/\Delta_2^2\phi_2}\vee  1} }{\phi_2}, \\
    & \Tilde{C}_2 = \frac{512(\log d +2)d x_{\max}^2 x_{\infty}^2 \sigma^2 \prns{1 \vee 64  x_{\max}^4 /\phi_2^2} }{ \Delta_1^2 \phi_1^2} ,
\end{align*}
we have that, if $q\geq \tilde{C}_1 \vee \tilde{C}_2$, then our expected regret is bounded by:
\begin{align}
R_T(\hat{\mathcal{A}}) \le &~ 8 q b_{\max} x_{\max} (3\log T + 128q) + 24 b_{\max} x_{\max}\prns{49+ \frac{16dx_{\max}^2}{\phi_1} + \frac{16d x_{\max}^2}{\phi_2}}\notag \\
    & ~ + 2^{\alpha_1+2}\gamma_1  C_1^{-\frac{\alpha_1+1}{2}} \prns{(M+2)^{\alpha_1+2}+ 2^{\alpha_1+6}}f(\alpha_1) \notag\\
    &~ + 2^{\alpha_2+3}\gamma_2  C_2^{-\frac{\alpha_2+1}{2}} \prns{(M+2)^{\alpha_2+2}+ 2^{\alpha_2+4}}f(\alpha_2), \label{eq:upperboundrate}
\end{align}
where
\begin{align*}
& f(\alpha') = \frac{2 T^{\frac{1-\alpha'}{2}}}{1-\alpha'}\ind\{\alpha'<1\} + \log T\ind\{\alpha' = 1\} + \frac{2}{\alpha'-1}\ind\{\alpha'>1\}, \\
 & C_1 = \min\braces{\frac{\phi_1^2 }{128 d x_\infty^2 x_{\max}^2 \sigma^2},\frac{\phi_1^2 \phi_2^2 }{2048 d x_\infty^2 x_{\max}^6 \sigma^2} },   \\
    & C_2 = \frac{\phi_2^2}{32 d x_{\max}^2 x_{\infty}^2 \sigma^2},\\
   & M = \lfloor \sqrt{(\alpha+4)\log ((\alpha+4) d)}\rfloor.
\end{align*}
\end{theorem}
{In the most relevant case of $\alpha = 1$ (\cf~\cref{lemma: alpha is one}), our algorithm achieves logarithmic regret, which significantly improves on the existing square-root regret (see \cref{section: literature}).}

{The regret rate obtained in \cref{eq:upperboundrate} is optimal in its dependence on $T$ for $\alpha\leq1$ and optimal up to a log factor for $\alpha>1$.
To see this first note that any one-stage linear-response contextual bandit instance $\prns{\bar{\mathcal{P}}_\bx, \bar{\bb}_1, \bar{\bb}_2, \bar{\mathcal{P}}_{\eta^{(1)}}, \bar{\mathcal{P}}_{\eta^{(2)}}}$ can be embedded inside an instance of the DTR bandit problem by simply letting $\bx_{t,1} \sim \bar{\mathcal{P}}_\bx$, $\bx_{t,2}(a_1) = \bar{\bb}_{a_1}^\intercal \bx_{t,1} + \eta_t^{(a_1)}$, $\bx_{t,3}(a_1, a_2) \equiv 0$, and $Y_t = \bx_{t,2}$.
In this construction, we essentially observe the same information and receive the same reward in the DTR bandit problem as in the one-stage bandit problem.
Conversely, given any DTR bandit algorithm $\mathcal{A}$, we can apply the algorithm in the one-stage linear-response bandit problem by treating it as a DTR instance with $\bx_{t,1} \sim \bar{\mathcal{P}}_\bx$, $\bx_{t,2}(a_1) = \bar{\bb}_{a_1}^\intercal \bx_{t,1} + \eta_t^{(a_1)}$, $\bx_{t,3}(a_1, a_2) \equiv 0$, and $Y_t = \bx_{t,2}$.
And, the regret of $\mathcal{A}$ up to round $T$ in the DTR bandit problem instance constructed above is exactly equal to its regret up to round $T$ in the given one-stage linear-response bandit instance.
}

\citet{bastani2021mostly} show that, in the one-stage linear-response contextual bandit problem, for \emph{any} algorithm, there always exists an instance satisfying the first-stage parts of our assumptions (namely, \cref{assumption: bounded} for the first-stage parameters, \cref{eq: margin 1} of \cref{assumption: margin}, and \cref{assumption3: 1} of \cref{assumption: diversity}) such that
the regret up to time $T$ is $\Omega\prns{T^{\frac{1-\alpha}{2}}}$ when $\alpha<1$, $\Omega(\log T)$ when $\alpha = 1$, and $\Omega(1)$ when $\alpha>1$,
where the asymptotic notation omits constants in $T$.
By our above observation, such an instance can also be embedded into a DTR bandit instance with $\alpha_1 = \alpha$ and $\alpha_2 = \infty$,
so the same lower bound applies to algorithms for the DTR bandit problem.
Thus, \cref{alg} is rate-optimal in $T$ when $\alpha\le 1$, and only exceeds the lower bound by a factor of $\log T$ when $\alpha>1$.

\section{Empirical Study} \label{section: empirics}
We now compare the performance of DTRBandit with some benchmark algorithms on both synthetic data and the STAR*D dataset obtained from a randomized clinical trial.

\subsection{Algorithms Considered}\label{section: numerics algorithms}
For both synthetic and STAR*D experiments, we will compare our algorithm with the following online benchmarks:
{
\begin{enumerate}
    \item $\epsilon$-Greedy: At the beginning, we pull each of $(a_1, a_2)$ pairs $d$ times to enable initial estimation for $\hat{\bb}_{a,m}$. In all subsequent rounds, with probability $1-\epsilon$ we act greedily according to the $\hat{Q}$ estimates, and with probability $\epsilon$ we pull the other arm to explore. Greedy is a special case with $\epsilon = 0$.
    \item LSVI-UCB: algorithm 1 in \cite{jin2019provably}, an upper-confidence-bound-based algorithm designed for linear MDPs. The tuning parameters are chosen as suggested in theorem $3.1$ therein.
\end{enumerate}
}

\text{DTRBandit} depends on the tuning parameters $\Delta_1$, $\Delta_2$, and $q$. We will investigate the how the algorithms performance with different parameter choices. 
{In all experiments, we set $\Delta_1=\Delta_2=\Delta$ and abbreviate the corresponding algorithm as DTRB$(q, \Delta)$ .}

\begin{figure}[t!]\centering%
\includegraphics[width = 8cm]{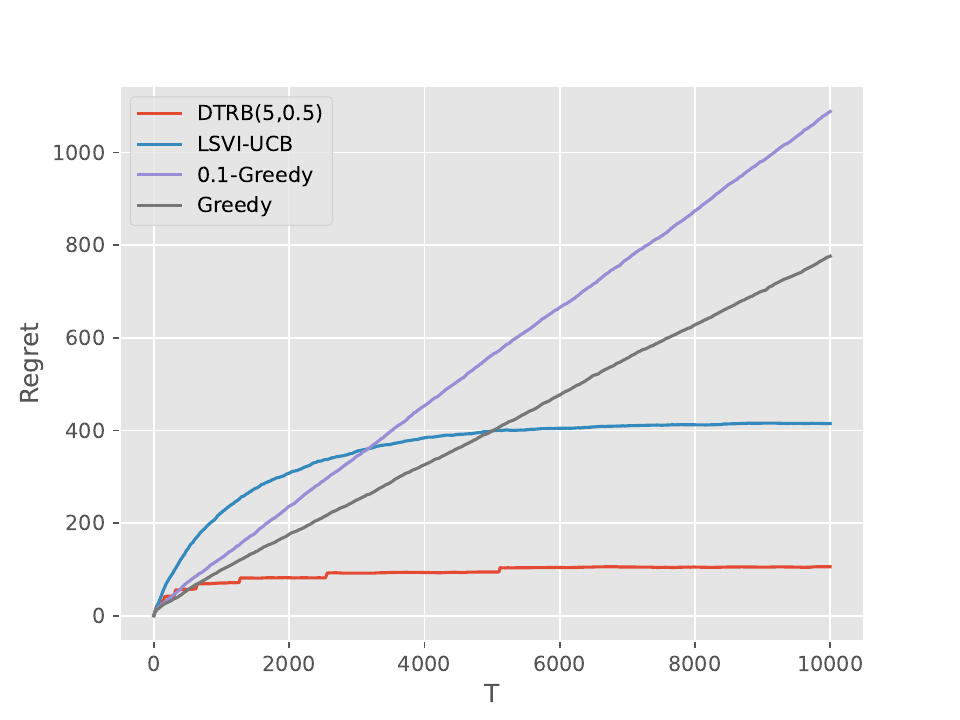}
\caption{Expected regret of different algorithms. }
\label{fig: different algs}
\end{figure}

\subsection{Synthetic Experiments} \label{section: synthetic experiments}
{ 

We start by considering a synthetic experiment where we generate data from a model that conforms to our linear $Q$ assumption.
We let $\bx_{t,1}$ be i.i.d samples from $\text{Uniform}(0,1)$. The second-stage contexts are generated from the following distributions:
\begin{align*}
    \pr\prns{\bx_{t,2}(1) = \textbf{x}_2 \mid \bx_{t,1} = \textbf{x}_1 } = \textbf{x}_1\text{Beta}_{4,1}(\textbf{x}_2) + (1 - \textbf{x}_1)\text{Beta}_{1,2}(\textbf{x}_2), \\
    \pr\prns{\bx_{t,2}(2) = \textbf{x}_2 \mid \bx_{t,1} = \textbf{x}_1 } =  \textbf{x}_1\text{Beta}_{1,4}(\textbf{x}_2) + (1 - \textbf{x}_1)\text{Beta}_{2,1}(\textbf{x}_2).
\end{align*}
Lastly, $\bx_{t,3}(a_1, 1) = \bx_{t,2}(a_1) + 1 +\epsilon_t^{(a_1, 1)} $, and $\bx_{t,3}(a_1, 2) = 3\bx_{t,2}(a_1) + \epsilon_t^{(a_1, 2)}$, where each $\epsilon_t^{(a_1, a_2)}$ is i.i.d. from $\text{Normal}(0,1)$.
The final reward $Y(a_1, a_2)$ equals $\bx_{t,3}(a_1, a_2)$.
With these distributions, we can compute the $Q$-functions in closed form, and the corresponding optimal decision rule is: in the first stage, pull arm $1$ when $\bx_1 > 5/14$ and arm $2$ otherwise; in the second stage, pull arm $1$ when $\bx_2 < 1/2$ and arm $2$ otherwise.
It is easy to check that our example satisfy the conditions in \cref{lemma: alpha is one}, so it satisfies \cref{assumption: margin} with $\alpha = 1$.

\cref{fig: different algs} shows the regret for different algorithms in this setting. Here we simulate $40$ independent paths up to $T=10{,}000$, and compute the average regret of different algorithms over these paths. 
Both Greedy and $\epsilon$-Greedy achieve linear regret in the long run, but for different reasons.
Greedy suffers from a constant fraction of ``bad" paths where the decision-maker stops exploring and repeatedly makes the suboptimal decision from early on due to lack of exploration.
On the other hand, $\epsilon$-Greedy suffers from excessive exploration: randomizing with $\epsilon$ probability at each time step means it chooses the sub-optimal arm with $\Omega(\epsilon)$ probability at each time step, which accumulates to $\Omega(\epsilon T)$ regret.
LSVI-UCB achieves sub-linear regret, but performs significantly worse than DTRBandit.
In contrast to these three, \text{DTRBandit} performs very well in the long run and has logarithmic regret.

\begin{figure}[t!]\centering%
\begin{subfigure}[t]{0.33\textwidth}\includegraphics[width=\textwidth]{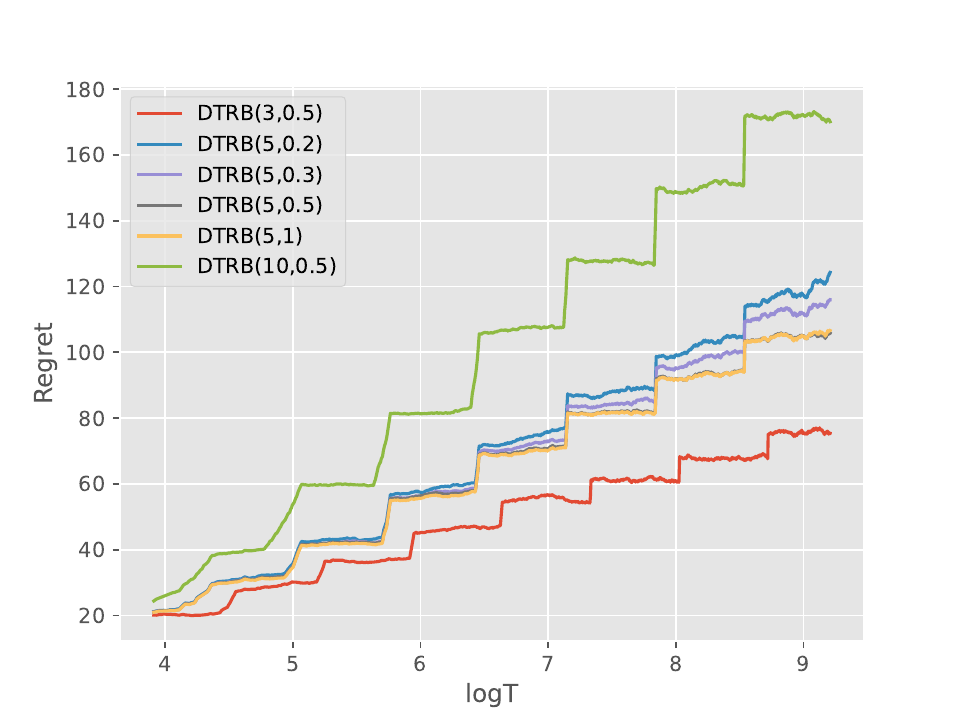}\caption{Regret versus $\log T$.} \label{subfig: dtr rates}\end{subfigure}%
\hfill\begin{subfigure}[t]{0.33\textwidth}\includegraphics[width=\textwidth]{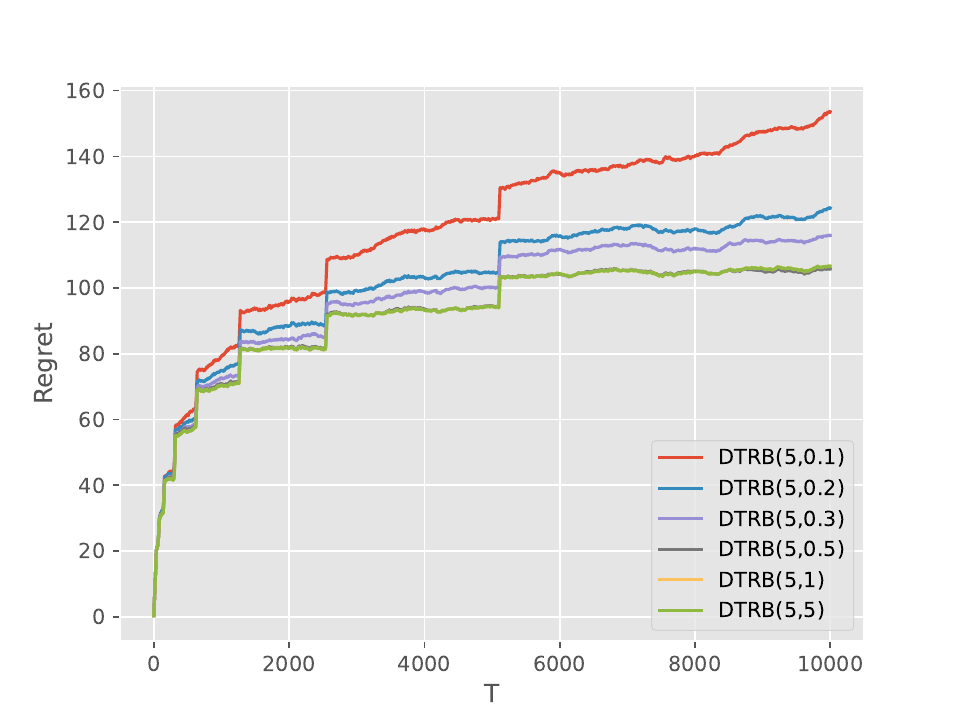}\caption{Regret with different $\Delta$'s.}\end{subfigure}%
\hfill\begin{subfigure}[t]{0.33\textwidth}\includegraphics[width=\textwidth]{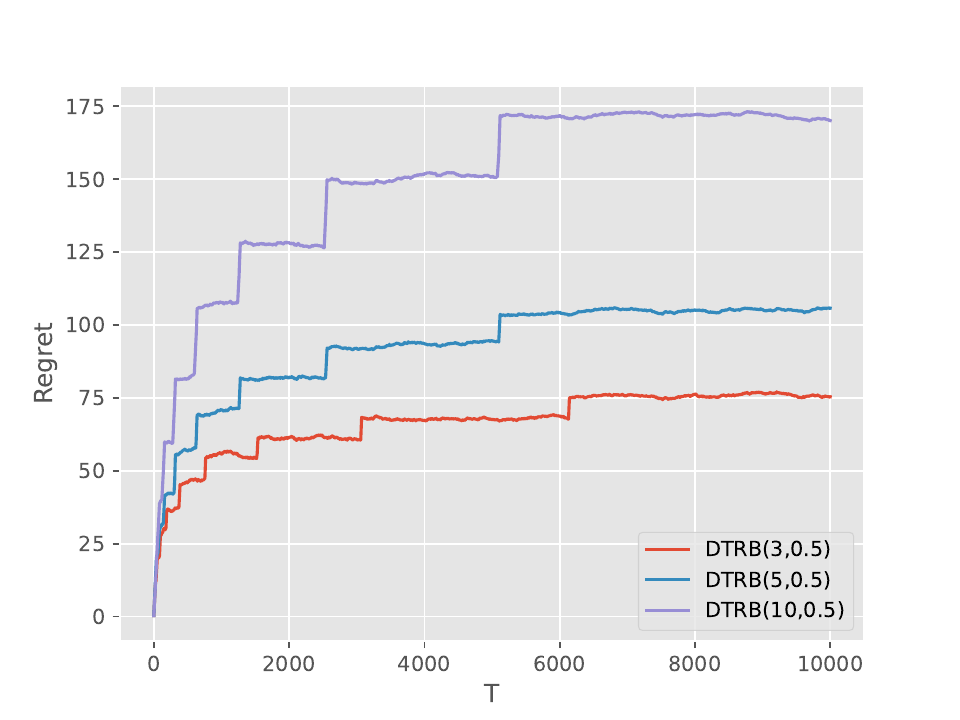}\caption{Regret with different $q$'s.}\end{subfigure}
\caption{Expected regret of \textbf{DTRBandit} with different $\Delta$ and $q$ inputs.}
\label{fig: simulations}
\end{figure}

We now briefly discuss the choice of tuning parameters $\Delta$ and $q$ for DTRBandit. 
On a high level, larger $q$ means we perform more exploration, and larger $\Delta$ means we rely more on hat-estimators instead of tilde-estimators.
\cref{fig: simulations} shows the expected regret of DTRBandit with different $\Delta$ and $q$ inputs.
First of all, DTRBandit is very robust to the choices of these parameters, and \cref{subfig: dtr rates} shows that it achieves logarithmic regret across different parameter choices, which agrees with our theoretical results in \cref{section: theory}.
It is interesting to see that we can achieve logarithmic regret in practice even when $q$ is much smaller than what would be theoretically required by \cref{thm: regret upper bound}, and $\Delta$ is much larger than what would be theoretically required to satisfy \cref{assumption: diversity}.
Moreover, the cumulative regret gets lower as $q$ gets smaller, and $\Delta$ gets larger.
Therefore, in practice, we suggest following a ``small $q$, large $\Delta$ (compared to the scale of rewards)" rule-of-thumb to choose these parameters.

}

\subsection{Empirical Study: STAR*D Data} \label{section: stard}

We next study the algorithms using real data, where 
our {linear $Q$ assumption} likely does not hold.
\subsubsection{Background on STAR*D and Problem Formulation.}
Sequenced Treatment Alternatives to Relieve Depression (STAR*D) was a multi-level randomized controlled trial designed to assess effectiveness of different treatment regimes for patients with major depressive disorder \citep{rush2004sequenced}.
The primary outcome at each level (\ie, stage) was assessed using the QIDS scores, which measures severity of depression and ranges from $0$ to $26$ in the sample.
Before each treatment level, participants with a total clinican-rated QIDS-score of $0$ to $5$ were considered as having a clinically meaningful response and therefore in remission, and those without remission were eligible for future treatments and entered the next treatment level.

Following existing literature \citep{chakraborty2013inference,liu2018augmented}, we focus on level $2/2A$ (referred to as stage $1$) and level $3$ (referred to as stage $2$) of the study only.
There were $1260$ participants with complete features who entered stage $1$; $468$ of them achieved remission, $465$ of them dropped out, and $327$ of them entered stage $2$.
The treatments were divided into two groups: one group that involves SSRIs (selective serotonin reuptake inhibitors), %
and the other that involves only non-SSRIs.

We define the following context, treatment, and outcome variables:
\begin{itemize}
    \item $\bx_1$: QIDS-score measured at the start of stage $1$ (QIDS.start$_1$), the slope of QIDS-score over the previous level (QIDS.slope$_1$), and participant preference at the start of stage $1$ (preference$_1$, taking values $1,2$). We let $\psi_1(\bh_1) = \bx_1$.

    \item $A_1$: $1$ if treated with SSRI drugs in stage $1$, and $2$ if treated with non-SSRI drugs.

    \item $\bx_2$: QIDS.end$_1$ measured at the start of stage $1$, QIDS.start$_2$, QIDS.slope$_2$ and preference$_2$ measured at the start of stage $2$. We let $\psi_2(\bh_2) = (\text{QIDS.start}_2, \text{QIDS.slope}_2$, $\text{preference}_2)$.

    \item {$A_2$: $1$ if treated with SSRI drugs in stage $2$, $2$ if treated with non-SSRI drugs. In the dataset, if a patient is treated with a non-SSRI drug in stage $1$, then $A_2 \equiv 2$.}
    
    \item $\bx_3$: QIDS.end$_2$ measured at the start of stage $2$.

    \item { $Y$: Let $R=1$ if the patient achieves remission before stage $2$, and $0$ otherwise. Define
    \begin{align*}
        Y = -R\cdot \text{QIDS.end}_1 - (1-R)\cdot \prns{\frac{ \text{QIDS.end}_1 +  \text{QIDS.end}_2}{2}}.
    \end{align*}
    Here we take the negative of QIDS-scores to make higher values correspond to better outcomes.}
\end{itemize}

The STAR*D dataset is then viewed as 1260 observations of the above variables, one for each participant with complete features who entered stage 1.
However, in contrast to existing literature that focuses on \textit{offline} DTR estimation and evaluation methods, we aim to learn a mapping between patient covariates and the optimal treatment regime in a \textit{online} and \textit{adaptive} manner. Therefore, we do not simply observe all data points at once nor do we even observe all of them sequentially. We next discuss how we use this data to nonetheless evaluate online DTR bandit algorithms.

\subsubsection{Off-policy DTR Bandit Algorithm Evaluation}
For any given DTR bandit algorithm $\mathcal{A}$, we want to evaluate its performance using the average reward it collects up to each fixed time $T$, \ie,
$${g}_T(\mathcal{A}) = \expect\bracks{\frac{1}{T}\sum_{t=1}^T Y_t(A_{t,1}, A_{t,2}) },$$
where $A_{t,1}$ and $A_{t,2}$ are obtained by running algorithm $\mathcal{A}$.
Because we only have access to outcomes of the treatments administered in the STAR*D dataset and never for treatments not administered, estimating ${g}_T(\mathcal{A})$ using this dataset is nontrivial. This is known as off-policy evaluation.

Our key idea for off-policy DTR bandit algorithm evaluation stems from the Policy-Evaluator Algorithm \citep[Algorithm $1$]{li2011unbiased}, which gives an unbiased estimator of the average reward for contextual bandit algorithms.
Assume that $S$ is a dataset obtained from a \textit{uniformly} randomized sequential trial, \ie, $(a_1,a_2)$ are chosen uniformly at random to generate the data. Then we could evaluate a DTR bandit algorithm $\mathcal A$ as follows. We play through the data sequentially.
First we set $t=0$.
For each data point $(\textbf{x}_1, a_1, \textbf{x}_2, a_2, \textbf{x}_3, y)$ in $S$, we first show $\textbf{x}_1$ to $\mathcal A$. If it takes an action different from $a_1$ in the first stage, we skip this data point, ignoring the outcomes, not adding it to our history, and not increasing the time counter $t$. Otherwise, we additionally show $(a_1,\textbf{x}_2)$ to $\mathcal A$. Again, if it takes an action different from $a_2$ in the second stage, we skip this data point. Otherwise, if it matched both $a_1$ and $a_2$, we add the outcome (divided by $T$) to our cumulative reward, add the data point to our history, and increment $t$.
We repeat this procedure until we get $T$ samples in our trajectory.
In the end, we must have an unbiased estimator for ${g}_T(\mathcal{A})$.

\begin{algorithm}[t!]%
    \caption{\textsc{ReplayDTR}}
    \label{alg: replay}
    \begin{algorithmic}[1]
    \STATE \textbf{Inputs:} $T > 0$; online DTR algorithm $\mathcal{A}$; dataset $S$; weights $\{1/p_i\}_{i\in S}$.
    \STATE $\text{hist}_0 \leftarrow ()$ \quad \{An initially empty history\}
    \STATE $\hat{G}(\mathcal{A}) \leftarrow 0 $ \quad \{An initially zero total payoff\}
    \FOR{$t = 1,2,\dots, T$}
        \STATE \textbf{repeat}
        \STATE \quad {Sample the next data point $(\mathbf{x}_1, a_1,\mathbf{x}_2, a_2, \mathbf{x}_3, y)$ according to weights $\{1/p_i\}_{i\in S}$ from $S$ with replacement}
        \STATE \textbf{until} $\mathcal{A}(\text{hist}_{t-1}, \mathbf{x}_1) = a_1,\, \mathcal{A}(\text{hist}_{t-1},\mathbf{x}_1, a_1, \mathbf{x}_2) = a_2$
        \STATE $\text{hist}_t \leftarrow (\text{hist}_{t-1}, \mathbf{x}_1, a_1, \mathbf{x}_2, a_2, \mathbf{x}_3, y)$
        \STATE $\hat{G}(\mathcal{A}) \leftarrow \hat{G}(\mathcal{A}) + y$
    \ENDFOR
    \item Output: $\hat{G}(\mathcal{A})/T$
    \end{algorithmic}
\end{algorithm}

However, the STAR*D data does \emph{not} come from a uniformly randomized trial and the randomization probabilities are adapted to each participant's time-varying features (but not adapted over participants).
Moreover, since we are evaluating a history-dependent online algorithm rather than a fixed policy, we cannot simply reweight the outcomes in the average reward computed from the previously mentioned method.
{We tackle this problem by using weighted sampling from STAR*D data (with replacement), so that each data point looks like coming from uniform randomization between arms. }
Specifically,
first, we estimate the propensity $p_i$ of the $i$-th point $(\textbf{x}_{1i}, a_{1i}, \textbf{x}_{2i}, a_{2i}, \textbf{x}_{3i}, y_{i})$ in STAR*D, \ie, the probability we choose $(a_{1i}, a_{2i})$ given the features. $p_i$ is estimated as the product of three parts:
i) the probability of choosing $a_{1i}$ in stage $1$ (from a logistic model using $\textbf{x}_{1i}$ as predictors),
ii) the probability of stay/drop-out in stage $2$ (from a logistic model using $\textbf{x}_{1i}$, $a_{1i}$, $\textbf{x}_{1i}a_{1i}$ and QIDS.slope$_2$ as predictors), and
iii) the probability of choosing $a_{i,2}$ in stage $2$ if the $i$-th point is present in stage $2$ (from a logistic model using $\textbf{x}_{1i}$, $a_{1i}$, $\textbf{x}_{1i}a_{1i}$ and $\textbf{x}_{2i}$ as predictors).
{We then assign weight $1/p_i$ to the $i$-th data point, and to generate the next data point $(\mathbf{x}_1, a_1,\mathbf{x}_2, a_2, \mathbf{x}_3, y)$, we sample from $S$ according to weights $\{1/p_i\}_{i\in S}$ with replacement.
The rest of our algorithm is the same as in the uniformly randomized case, and the detailed evaluation procedure is summarized in \cref{alg: replay}.}

\subsubsection{Comparison of Algorithms}
Again, we compare the empirical performance of DTRBandit with Greedy, $\epsilon$-Greedy and LSVI-UCB.
In addition, we use the following two as benchmarks for comparison:
\begin{enumerate}
    \item Logging: the average rewards observed in the STAR*D dataset after adjusting for dropouts (which estimates $g_T(\mathcal A)$ for the logging policy that generated the data).
    \item Offline: the average rewards of the estimated optimal treatment regime using the whole offline dataset (given in \citealp{chakraborty2013inference}), which suggests treating a patient in stage $1$ with SSRI if $(-0.73+0.01\times\text{QIDS.start}_1 + 0.01\times \text{QIDS.slope}_1-0.67\times\text{preference}_1)>0$ (and with a non-SSRI otherwise), and treating a patient in stage $2$ with SSRI if $(-0.18-0.01\times \text{QIDS.start}_2 -0.25\times \text{QIDS.slope}_2)>0$ (and with a non-SSRI otherwise).
\end{enumerate}

{

To be consistent with real-world practice, all algorithms are implemented with the extra condition that we do not enter the second stage when $\text{QIDS.end}_1 \le 5$, and we can only use non-SSRI drugs in stage $2$ if we use non-SSRI drugs in stage $1$.
When implementing DTRBandit and $\epsilon$-Greedy, because when $\text{QIDS.end}_1 \le 5$, the outcome $Y$ is determined by the observations before stage $2$, we define $\tilde{Y}= \hat{Y} = \text{QIDS.end}_1 $ for data points with $\text{QIDS.end}_1 \le 5$; we attempted a similar change when implementing LSVI-UCB, but its performance gets worse, so we do not make this change for LSVI-UCB.
Besides, for all algorithms, in both defining and estimating $Q_1(\textbf{h}, 2)$, we replace the max with the estimated value of pulling arm $2$ in stage $2$, because arm $2$ is the only feasible arm afterwards. 
Our exploration schedule for DTRBandit is updated to explore the $3$ first- and second-stage treatment combinations that appear in the data, because we do not need to learn about the infeasible treatment plan $(2, 1)$. %
Moreover, since \cite{jin2019provably} require the feature map to be bounded by $1$, we normalize the features when running LSVI-UCB.
Lastly, for \text{DTRBandit}, we vary the margin parameter $\Delta$ and force-pull schedule parameter $q$, while following the general rule of ``small $q$, large $\Delta$" as we discussed in \cref{section: synthetic experiments}. Compared with synthetic experiments, we choose smaller $q$'s since the horizon is shorter, and larger $\Delta$'s since the scale of the final outcome is much larger.

\begin{table}\footnotesize
\newcolumntype{A}{>{$}p{1.2cm}<{$}}
\newcolumntype{B}{>{\hspace{-.4cm}$}p{.6cm}<{$}}
\newcolumntype{C}{>{\centering\arraybackslash$}c<{$}}
\setlength{\tabcolsep}{8pt}
\centering
\caption{Average rewards of different algorithms at $T=5000$}
\label{table: stard}
\begin{tabular}{CCCCCCC}
 \toprule
 
\text{DTRB}$(1, 10)$  &  \text{DTRB}$(1, 20)$ & \text{DTRB}$(1, 50)$ &  \text{DTRB}$(2, 10)$ & \text{DTRB} $(2, 20)$  &  \text{DTRB} $(2, 50)$\\
 -7.680 &  -7.561 & -7.552 &  -7.688  & -7.604  &-7.547 \\
 \midrule
 \text{Logging} & \text{LSVI-UCB} & \text{Greedy} &  $0.1$\text{-Greedy} & $0.2$\text{-Greedy} & \text{Offline}  \\
 -8.438  & -8.071 &  -7.810 & -7.724  & -7.835  & -7.688 \\
 \bottomrule
\end{tabular}
\end{table}

\cref{table: stard} presents the average rewards evaluated at $T=5{,}000$. Here we
generated $10$ independent paths from the STAR*D data set, and compute the mean of the average rewards of different algorithms over these paths.
All online algorithms improve substantially from the logging policy used in the original dataset.
In particular, \text{DTRBandit} consistently performs well under different parameter choices, achieves the highest average reward among all online algorithms, and is often better than the Offline benchmark that roughly serves as a proxy for ``good algorithms".
Interestingly, we find that the performance of DTRBandit gets better when $q$ gets smaller and $\Delta$ gets larger. This is consistent with our discovery in \cref{section: synthetic experiments}, and can serve as a general guideline for choosing these parameters in practice.

}

\section{Conclusions and Future Directions} \label{section: conclusion}
In this paper, we define and solve the DTR bandit problem, where we can both personalize initial decision to each incoming unit \emph{and} adapt a second-linear decision after observing the effect of the first-line decision.
We propose a novel algorithm that achieves rate-optimal regret under the linear $Q$ model.
Our algorithm shows favorable empirical performance in both synthetic scenarios and in a real-world healthcare dataset.

{
We discuss several extensions of work, some of which we will detail in the appendix. 
In \cref{section: delay}, we extend our model to the setting where the feedback from second-stage outcomes may be delayed until after a new unit already arrives. We provide a revised algorithm for a large class of arrival processes, and show that the additional regret due to delays is additive.
In \cref{section: k arms}, we discuss the extension to two-stage DTR bandits with more than two arms, where we also allow for the possibility of arms that are everywhere suboptimal, in violation of \cref{assumption: diversity}. We extend our analysis to show that a slightly modified algorithm has the same order of regret as in \cref{thm: regret upper bound}.
Finally, the extension to $M>2$ stages is straightforward. For each intermediate stage $m \in [M-1]$, we can compute $\tilde{Y}_{m,j} = \max_{a\in[2]} \tilde{\bb}_{a, m+1}^\intercal(t) \bp_{m+1}(\bh_{j,m+1})$ and $\hat{Y}_{m,j} = \max_{a\in[2]} \hat{\bb}_{a, m+1}^\intercal(t) \bp_{m+1}(\bh_{j,m+1})$. The corresponding $\tilde{\bb}_{a,m}(t)$ and $\hat{\bb}_{a,m}(t)$ can be obtained by regressing $\tilde{Y}_{m,j}$ and $\hat{Y}_{m,j}$ on $\bp_m(\bh_{j, m})$. The rest of the algorithm can be extended by changing $m\in [2]$ to $m\in [M]$.
}

\bibliographystyle{plainnat}
\bibliography{literature}

\begin{thebibliography}{44}
\providecommand{\natexlab}[1]{#1}
\providecommand{\url}[1]{\texttt{#1}}
\expandafter\ifx\csname urlstyle\endcsname\relax
  \providecommand{\doi}[1]{doi: #1}\else
  \providecommand{\doi}{doi: \begingroup \urlstyle{rm}\Url}\fi

\bibitem[Agarwal et~al.(2020)Agarwal, Kakade, Krishnamurthy, and
  Sun]{agarwal2020flambe}
Alekh Agarwal, Sham Kakade, Akshay Krishnamurthy, and Wen Sun.
\newblock Flambe: Structural complexity and representation learning of low rank
  mdps.
\newblock \emph{Advances in neural information processing systems},
  33:\penalty0 20095--20107, 2020.

\bibitem[Athey and Wager(2021)]{athey2021policy}
Susan Athey and Stefan Wager.
\newblock Policy learning with observational data.
\newblock \emph{Econometrica}, 89\penalty0 (1):\penalty0 133--161, 2021.

\bibitem[Athey et~al.(2018)Athey, Baird, Jamison, McIntosh, \"Ozler, and
  Sama]{atheytrial}
Susan Athey, Sarah Baird, Julian Jamison, Craig McIntosh, Berk \"Ozler, and
  Dohbit Sama.
\newblock A sequential and adaptive experiment to increase the uptake of
  long-acting reversible contraceptives in cameroon, 2018.
\newblock URL
  \url{http://pubdocs.worldbank.org/en/606341582906195532/Study-Protocol-Adaptive-experiment-on-FP-counseling-and-uptake-of-MCs.pdf}.
\newblock Study protocol.

\bibitem[Auer(2002)]{auer2002using}
Peter Auer.
\newblock Using confidence bounds for exploitation-exploration trade-offs.
\newblock \emph{Journal of Machine Learning Research}, 3\penalty0
  (Nov):\penalty0 397--422, 2002.

\bibitem[Azar et~al.(2017)Azar, Osband, and Munos]{azar2017minimax}
Mohammad~Gheshlaghi Azar, Ian Osband, and R{\'e}mi Munos.
\newblock Minimax regret bounds for reinforcement learning.
\newblock In \emph{Proceedings of the 34th International Conference on Machine
  Learning-Volume 70}, pages 263--272. JMLR. org, 2017.

\bibitem[Bastani and Bayati(2020)]{bastani2020online}
Hamsa Bastani and Mohsen Bayati.
\newblock Online decision making with high-dimensional covariates.
\newblock \emph{Operations Research}, 68\penalty0 (1):\penalty0 276--294, 2020.

\bibitem[Bastani et~al.(2021)Bastani, Bayati, and Khosravi]{bastani2021mostly}
Hamsa Bastani, Mohsen Bayati, and Khashayar Khosravi.
\newblock Mostly exploration-free algorithms for contextual bandits.
\newblock \emph{Management Science}, 67\penalty0 (3):\penalty0 1329--1349,
  2021.

\bibitem[Beygelzimer et~al.(2011)Beygelzimer, Langford, Li, Reyzin, and
  Schapire]{beygelzimer2011contextual}
Alina Beygelzimer, John Langford, Lihong Li, Lev Reyzin, and Robert Schapire.
\newblock Contextual bandit algorithms with supervised learning guarantees.
\newblock In \emph{Proceedings of the Fourteenth International Conference on
  Artificial Intelligence and Statistics}, pages 19--26, 2011.

\bibitem[Chakraborty and Moodie(2013)]{chakraborty2013statistical}
Bibhas Chakraborty and Erica~E.M. Moodie.
\newblock \emph{Statistical methods for dynamic treatment regimes}.
\newblock Springer, 2013.

\bibitem[Chakraborty et~al.(2010)Chakraborty, Murphy, and
  Strecher]{chakraborty2010inference}
Bibhas Chakraborty, Susan Murphy, and Victor Strecher.
\newblock Inference for non-regular parameters in optimal dynamic treatment
  regimes.
\newblock \emph{Statistical methods in medical research}, 19\penalty0
  (3):\penalty0 317--343, 2010.

\bibitem[Chakraborty et~al.(2013)Chakraborty, Laber, and
  Zhao]{chakraborty2013inference}
Bibhas Chakraborty, Eric~B Laber, and Yingqi Zhao.
\newblock Inference for optimal dynamic treatment regimes using an adaptive
  m-out-of-n bootstrap scheme.
\newblock \emph{Biometrics}, 69\penalty0 (3):\penalty0 714--723, 2013.

\bibitem[Dudik et~al.(2011)Dudik, Hsu, Kale, Karampatziakis, Langford, Reyzin,
  and Zhang]{dudik2011efficient}
Miroslav Dudik, Daniel Hsu, Satyen Kale, Nikos Karampatziakis, John Langford,
  Lev Reyzin, and Tong Zhang.
\newblock Efficient optimal learning for contextual bandits.
\newblock In \emph{Proceedings of the Twenty-Seventh Conference on Uncertainty
  in Artificial Intelligence}, pages 169--178, 2011.

\bibitem[Goldenshluger and Zeevi(2013)]{goldenshluger2013linear}
Alexander Goldenshluger and Assaf Zeevi.
\newblock A linear response bandit problem.
\newblock \emph{Stochastic Systems}, 3\penalty0 (1):\penalty0 230--261, 2013.

\bibitem[Hu et~al.(2022)Hu, Kallus, and Mao]{hu2022smooth}
Yichun Hu, Nathan Kallus, and Xiaojie Mao.
\newblock Smooth contextual bandits: Bridging the parametric and
  nondifferentiable regret regimes.
\newblock \emph{Operations Research}, 2022.

\bibitem[Jiang et~al.(2019)Jiang, Song, Li, and Zeng]{jiang2019entropy}
Binyan Jiang, Rui Song, Jialiang Li, and Donglin Zeng.
\newblock Entropy learning for dynamic treatment regimes.
\newblock \emph{Statistica Sinica}, 29\penalty0 (4):\penalty0 1633, 2019.

\bibitem[Jin et~al.(2019)Jin, Yang, Wang, and Jordan]{jin2019provably}
Chi Jin, Zhuoran Yang, Zhaoran Wang, and Michael~I Jordan.
\newblock Provably efficient reinforcement learning with linear function
  approximation.
\newblock \emph{arXiv preprint arXiv:1907.05388}, 2019.

\bibitem[Kallus(2018)]{kallus2018balanced}
Nathan Kallus.
\newblock Balanced policy evaluation and learning.
\newblock In \emph{Advances in Neural Information Processing Systems}, pages
  8895--8906, 2018.

\bibitem[Lai and Robbins(1985)]{lai1985asymptotically}
Tze~Leung Lai and Herbert Robbins.
\newblock Asymptotically efficient adaptive allocation rules.
\newblock \emph{Advances in applied mathematics}, 6\penalty0 (1):\penalty0
  4--22, 1985.

\bibitem[Li et~al.(2011)Li, Chu, Langford, and Wang]{li2011unbiased}
Lihong Li, Wei Chu, John Langford, and Xuanhui Wang.
\newblock Unbiased offline evaluation of contextual-bandit-based news article
  recommendation algorithms.
\newblock In \emph{Proceedings of the fourth ACM international conference on
  Web search and data mining}, pages 297--306, 2011.

\bibitem[Liu et~al.(2018)Liu, Wang, Kosorok, Zhao, and Zeng]{liu2018augmented}
Ying Liu, Yuanjia Wang, Michael~R Kosorok, Yingqi Zhao, and Donglin Zeng.
\newblock Augmented outcome-weighted learning for estimating optimal dynamic
  treatment regimens.
\newblock \emph{Statistics in medicine}, 37\penalty0 (26):\penalty0 3776--3788,
  2018.

\bibitem[Mammen and Tsybakov(1999)]{mammen1999smooth}
Enno Mammen and Alexandre~B Tsybakov.
\newblock Smooth discrimination analysis.
\newblock \emph{The Annals of Statistics}, 27\penalty0 (6):\penalty0
  1808--1829, 1999.

\bibitem[Moodie et~al.(2014)Moodie, Dean, and Sun]{moodie2014q}
Erica~EM Moodie, Nema Dean, and Yue~Ru Sun.
\newblock Q-learning: Flexible learning about useful utilities.
\newblock \emph{Statistics in Biosciences}, 6\penalty0 (2):\penalty0 223--243,
  2014.

\bibitem[Murphy(2003)]{murphy2003optimal}
Susan~A Murphy.
\newblock Optimal dynamic treatment regimes.
\newblock \emph{Journal of the Royal Statistical Society: Series B (Statistical
  Methodology)}, 65\penalty0 (2):\penalty0 331--355, 2003.

\bibitem[Murphy et~al.(2001)Murphy, van~der Laan, and
  Robins]{murphy2001marginal}
Susan~A Murphy, Mark~J van~der Laan, and James~M Robins.
\newblock Marginal mean models for dynamic regimes.
\newblock \emph{Journal of the American Statistical Association}, 96\penalty0
  (456):\penalty0 1410--1423, 2001.

\bibitem[Pelham and Fabiano(2008)]{pelham2008evidence}
William~E Pelham and Gregory~A Fabiano.
\newblock Evidence-based psychosocial treatments for
  attention-deficit/hyperactivity disorder.
\newblock \emph{Journal of Clinical Child \& Adolescent Psychology},
  37\penalty0 (1):\penalty0 184--214, 2008.

\bibitem[Perchet and Rigollet(2013)]{perchet2013multi}
Vianney Perchet and Philippe Rigollet.
\newblock The multi-armed bandit problem with covariates.
\newblock \emph{The Annals of Statistics}, 41\penalty0 (2):\penalty0 693--721,
  2013.

\bibitem[Qian and Murphy(2011)]{qian2011performance}
Min Qian and Susan~A Murphy.
\newblock Performance guarantees for individualized treatment rules.
\newblock \emph{Annals of statistics}, 39\penalty0 (2):\penalty0 1180, 2011.

\bibitem[Qiang and Bayati(2016)]{qiang2016dynamic}
Sheng Qiang and Mohsen Bayati.
\newblock Dynamic pricing with demand covariates.
\newblock \emph{Available at SSRN 2765257}, 2016.

\bibitem[Rigollet and Zeevi(2010)]{rigollet2010nonparametric}
Philippe Rigollet and Assaf Zeevi.
\newblock Nonparametric bandits with covariates.
\newblock In Adam~Tauman Kalai and Mehryar Mohri, editors, \emph{{COLT} 2010 -
  The 23rd Conference on Learning Theory, Haifa, Israel, June 27-29, 2010},
  pages 54--66. Omnipress, 2010.

\bibitem[Rush et~al.(2004)Rush, Fava, Wisniewski, Lavori, Trivedi, Sackeim,
  Thase, Nierenberg, Quitkin, and Kashner]{rush2004sequenced}
A~John Rush, Maurizio Fava, Stephen~R Wisniewski, Philip~W Lavori, Madhukar~H
  Trivedi, Harold~A Sackeim, Michael~E Thase, Andrew~A Nierenberg, Frederic~M
  Quitkin, and T~Michael Kashner.
\newblock Sequenced treatment alternatives to relieve depression (star* d):
  rationale and design.
\newblock \emph{Controlled clinical trials}, 25\penalty0 (1):\penalty0
  119--142, 2004.

\bibitem[Schulte et~al.(2014)Schulte, Tsiatis, Laber, and
  Davidian]{schulte2014q}
Phillip~J Schulte, Anastasios~A Tsiatis, Eric~B Laber, and Marie Davidian.
\newblock Q-and a-learning methods for estimating optimal dynamic treatment
  regimes.
\newblock \emph{Statistical science: a review journal of the Institute of
  Mathematical Statistics}, 29\penalty0 (4):\penalty0 640, 2014.

\bibitem[Shortreed et~al.(2011)Shortreed, Laber, Lizotte, Stroup, Pineau, and
  Murphy]{shortreed2011informing}
Susan~M Shortreed, Eric Laber, Daniel~J Lizotte, T~Scott Stroup, Joelle Pineau,
  and Susan~A Murphy.
\newblock Informing sequential clinical decision-making through reinforcement
  learning: an empirical study.
\newblock \emph{Machine learning}, 84\penalty0 (1-2):\penalty0 109--136, 2011.

\bibitem[Song et~al.(2015)Song, Wang, Zeng, and Kosorok]{song2015penalized}
Rui Song, Weiwei Wang, Donglin Zeng, and Michael~R Kosorok.
\newblock Penalized q-learning for dynamic treatment regimens.
\newblock \emph{Statistica Sinica}, 25\penalty0 (3):\penalty0 901, 2015.

\bibitem[Sutton and Barto(2018)]{sutton2018reinforcement}
Richard~S Sutton and Andrew~G Barto.
\newblock \emph{Reinforcement learning: An introduction}.
\newblock MIT press, 2018.

\bibitem[Tewari and Murphy(2017)]{tewari2017ads}
Ambuj Tewari and Susan~A Murphy.
\newblock From ads to interventions: Contextual bandits in mobile health.
\newblock In \emph{Mobile Health}, pages 495--517. Springer, 2017.

\bibitem[Thall et~al.(2007)Thall, Wooten, Logothetis, Millikan, and
  Tannir]{thall2007bayesian}
Peter~F Thall, Leiko~H Wooten, Christopher~J Logothetis, Randall~E Millikan,
  and Nizar~M Tannir.
\newblock Bayesian and frequentist two-stage treatment strategies based on
  sequential failure times subject to interval censoring.
\newblock \emph{Statistics in medicine}, 26\penalty0 (26):\penalty0 4687--4702,
  2007.

\bibitem[Tropp(2015)]{tropp2015introduction}
Joel~A Tropp.
\newblock An introduction to matrix concentration inequalities.
\newblock \emph{Foundations and Trends{\textregistered} in Machine Learning},
  8\penalty0 (1-2):\penalty0 1--230, 2015.

\bibitem[Villar et~al.(2015)Villar, Bowden, and Wason]{villar2015multi}
Sof{\'\i}a~S Villar, Jack Bowden, and James Wason.
\newblock Multi-armed bandit models for the optimal design of clinical trials:
  benefits and challenges.
\newblock \emph{Statistical science}, 30\penalty0 (2):\penalty0 199, 2015.

\bibitem[Wainwright(2019)]{wainwright2019high}
Martin~J Wainwright.
\newblock \emph{High-dimensional statistics: A non-asymptotic viewpoint},
  volume~48.
\newblock Cambridge University Press, 2019.

\bibitem[Yang and Wang(2019)]{yang2019sample}
Lin Yang and Mengdi Wang.
\newblock Sample-optimal parametric q-learning using linearly additive
  features.
\newblock In \emph{International Conference on Machine Learning}, pages
  6995--7004. PMLR, 2019.

\bibitem[Zhang et~al.(2013)Zhang, Tsiatis, Laber, and
  Davidian]{zhang2013robust}
Baqun Zhang, Anastasios~A Tsiatis, Eric~B Laber, and Marie Davidian.
\newblock Robust estimation of optimal dynamic treatment regimes for sequential
  treatment decisions.
\newblock \emph{Biometrika}, 100\penalty0 (3):\penalty0 681--694, 2013.

\bibitem[Zhao et~al.(2015)Zhao, Zeng, Laber, and Kosorok]{zhao2015new}
Ying-Qi Zhao, Donglin Zeng, Eric~B Laber, and Michael~R Kosorok.
\newblock New statistical learning methods for estimating optimal dynamic
  treatment regimes.
\newblock \emph{Journal of the American Statistical Association}, 110\penalty0
  (510):\penalty0 583--598, 2015.

\bibitem[Zhao et~al.(2012)Zhao, Zeng, Rush, and Kosorok]{zhao2012estimating}
Yingqi Zhao, Donglin Zeng, A~John Rush, and Michael~R Kosorok.
\newblock Estimating individualized treatment rules using outcome weighted
  learning.
\newblock \emph{Journal of the American Statistical Association}, 107\penalty0
  (499):\penalty0 1106--1118, 2012.

\bibitem[Zhao et~al.(2009)Zhao, Kosorok, and Zeng]{zhao2009reinforcement}
Yufan Zhao, Michael~R Kosorok, and Donglin Zeng.
\newblock Reinforcement learning design for cancer clinical trials.
\newblock \emph{Statistics in medicine}, 28\penalty0 (26):\penalty0 3294--3315,
  2009.

\end{thebibliography}

\newpage

\appendix

\begin{center}
\LARGE\bf Appendices
\end{center}

\section{Regret Upper Bound Analysis}\label{sec:upperboundanalysis}

In this section, we provide an outline for the analysis of the regret upper bound. 
Here we break up our analysis into modular Propositions, each of which we prove in detail in \cref{section: proof}, and then we explain how these results come together to establish \cref{thm: regret upper bound}.

{

\subsection{OLS Tail Inequality for Adaptive Observations with Measurement Error.}\label{section: two tails}
The key to get uniform (over $\mathcal{H}_1$ or $\mathcal{H}_2$) concentration bounds on the estimated $Q$-functions is to get good concentration bounds for the OLS estimators $\Tilde\bb_{a,m},\hat\bb_{a,m}$.
However, in our bandit setup, the observations from pulling a certain arm are not independent.
Moreover, the true outcome $\max_{a_2\in[2]} Q_2\prns{\bh_{t,2}(a_1), a_2}$ is not observable in practice, and we can only run OLS on an imputed value with some possible measurement error.
In this section, we provide a general result for OLS estimators from adaptive samples with measurement error, which will be frequently utilized in later sections.

\begin{proposition}[OLS Tail Inequality for Adaptive Observations with Measurement Error]\label{prop: ols}
Let $S = \{(\bh_i, Y_i)\}_{i=1}^n$ be samples generated from a linear model $Y_i = \bb^T \mathbf{g}(\bh_i) + \epsilon_i + \delta_i$ with unknown $ \bb \in \mathbb{R}^d$ and known $\mathbf{g}: \mathcal{H}\rightarrow \mathbb{R}^d$.
Assume that there exists a filtration $\{\mathcal{F}_i\}_{i=0}^n$ such that $\{g_j(\bh_i) \epsilon_i\}_{i=1}^n$ is a martingale difference sequence adapted to the filtration, where $g_j(\bh_i)$ denotes the $j$-th entry of $\mathbf{g}(\bh_i)$.
If $||\mathbf{g}(\bh_i)|| \le x_{\max}$, $||\mathbf{g}(\bh_i)||_{\infty} \le x_{\infty}$ and $\epsilon_i$ is $\sigma$-sub-Gaussian for all $i\in [n]$ in an adaptive sense, the following tail inequality holds for all $\chi>0$ and $\phi>0$:
$$\pr\prns{\norm{\hat{\bb}-\bb}\ge \chi, \lambda_{\min}\prns{\hat{\Sigma}}\ge \phi}\le 2d\exp\prns{-\frac{\chi^2 \phi^2}{8n d x_{\infty}^2 \sigma^2}}+ \pr\prns{\sum_{i=1}^n\abs{ \delta_i }\ge \frac{\chi \phi}{2x_{\max}}}, $$
where $\hat{\Sigma} = \sum_{i=1}^n \mathbf{g}(\bh_i)\mathbf{g}^\intercal(\bh_i)$ and $\hat{\bb} = \hat{\Sigma}^{-1} \prns{\sum_{i=1}^n \mathbf{g}(\bh_i) Y_i}$.
\end{proposition}
}

\subsection{Concentration of the Forced-Pull Estimators.} \label{section: tilde estimators}
We are now equipped to prove the convergence of our forced-pull estimators. Define $\mathcal{G}_{t,m}$ to be the event that our $\Tilde{Q}_{t,m}$ has $\textit{good}$ estimation accuracy $\textit{uniformly}$ over both arms and all $\textbf{h}\in \mathcal{H}_m$:
\begin{equation}\label{eq: event gt}
 \mathcal{G}_{t,m} \triangleq \braces{\max_{a\in [2]}\sup_{\textbf{h}\in \mathcal{H}_m} |\Tilde{Q}_{t,m}(\textbf{h},a)-Q_m(\textbf{h},a)| \le \frac{\Delta_m}{4} }, \quad t\in [T], m\in [2],
\end{equation}
and $\mathcal{G}_{t}$ to be the event that both $\mathcal{G}_{t,1}$ and $\mathcal{G}_{t,2}$ hold at time $t$:
$$\mathcal{G}_t \triangleq \mathcal{G}_{t,1}\cap \mathcal{G}_{t,2}.$$
The following proposition shows that both $\mathcal{G}_{t,1}$ and $\mathcal{G}_{t,2}$ (thus, also $\mathcal{G}_t$) hold with high probability.
\begin{proposition}\label{prop: convergence of Tilde Q}
{When $q\geq \tilde{C}_1 \vee \tilde{C}_2$ and $t\ge 64q^2$},
\begin{enumerate}
    \item $\mathbb{P}(\mathcal{G}_{t,1})\ge 1- \frac{{8}}{t^2} ,$ \label{eq: gt1}
    \item $\mathbb{P}(\mathcal{G}_{t,2})\ge 1- \frac{{6}}{t^2}.$ \label{eq: gt2}
\end{enumerate}
\end{proposition}
To prove \cref{prop: convergence of Tilde Q}, we note that by the construction of forced sampling, $\{(\bh_{j,2}, Y_j): j\in \mathcal{T}_{a,2}(t)\}$ is a set of independent samples. Therefore, we can use \cref{prop: ols} to get concentration bounds for $\Tilde{\bb}_{a,2}(t)$.
{
When $\Tilde{\bb}_{a,2}(t)$ is well-concentrated, $\tilde{Y}_j$ is close enough to $\max_{a\in[2]} \bb^\intercal_{a,2} \bp_2(\bh_{j,2})$, so applying \cref{prop: ols} again we get concentration bounds for $\Tilde{\bb}_{a,1}(t)$.}
The only missing part is the positive-definiteness of the design matrices. Because $\{\bh_{j,m}: j\in \mathcal{T}_{a,m}(t)\}$ are independent samples, by the matrix Chernoff inequality (see, \eg, \citealp{tropp2015introduction}, Theorem 5.1.1), the minimal eigenvalue of the corresponding design matrix concentrates around the minimal eigenvalue of its mean. \cref{assumption: diversity} guarantees that the latter is strictly positive, thus our design matrix has lower-bounded minimal eigenvalue with high probability (\cref{lemma: positive defnite 1,lemma: positive defnite 2}). \cref{prop: convergence of Tilde Q} then follows naturally.

\subsection{Concentration of the All-Samples Estimators.} \label{section: hat estimators}
Next, we consider tail bounds on the all-samples estimators.
Define $\mathcal{M}_{t,m}$ to be the event that the \emph{minimum} eigenvalues of both $\hat{\Sigma}_{1,m}(t)$ and $\hat{\Sigma}_{2,m}(t)$ grow at least linearly in $t$:
{
\begin{align}
   \mathcal{M}_{t,m} = \braces{\lambda_{\min}\prns{ \hat{\Sigma}_{1,m}(t)}\ge \frac{\phi_m t}{4}} \cap \braces{\lambda_{\min}\prns{\hat{\Sigma}_{2,m}(t)} \ge \frac{\phi_m t}{4}} , \label{eq: event M}
\end{align}
}
For notation brevity, we let $\mathcal{M}_{t} =  \mathcal{M}_{t,1}\cap  \mathcal{M}_{t,2}$. 
{
Consider the samples $\braces{\prns{ \bh_{j,2}, Y_j}}_{j\in\mathcal{S}_{a,2}(t)}$.
Let $\mathcal{F}_j$ be the $\sigma$-algebra generated by $\braces{\bh_{i,2}, A_{i,2}, Y_i}_{i\in[j]} \cup \bh_{j+1,2}$. If we denote the $k$-th entry of $\bp_2\prns{\bh_{j,2}}$ as $\bp_{2k}\prns{\bh_{j,2}}$, it is easy to see that $\braces{ \bp_{2k}(\bh_{j,2})\eta_{j,2}^{(A_{j,1}, a)} \ind\braces{A_{j,2} = a}}_{j=1}^T$ is a martingale difference sequence adapted to $\braces{\mathcal{F}_j}_{j=1}^T$.
Therefore, we can use similar arguments as in the proof of \cref{prop: ols} to get tail bounds on $\hat{\bb}_{a,1}(t)$ under the event $\mathcal{M}_t$.
Moreover, we can apply similar arguments to $\braces{\prns{ \bh_{j,1}, \hat{Y}_j}}_{j\in\mathcal{S}_{a,2}(t)}$ to get a concentration bound on $\hat{\bb}_{a,2}(t)$ under $\mathcal{M}_t$.
Combining all these parts together we get the following two results:
\begin{proposition}\label{prop: convergence of hat Q 1}
For any $t\in [T]$,
\begin{align*}
  \pr\prns{\max_{\textbf{h}\in \mathcal{H}_1} \abs{\hat{Q}_{t,1}(\textbf{h},a)-Q_1(\textbf{h},a)}> \chi, \mathcal{M}_{t}}  \le 2d\exp\prns{-\frac{\phi_1^2 \chi^2 t}{128 d x_\infty^2 x_{\max}^2 \sigma^2}}  + 4d \exp\prns{-\frac{\phi_1^2 \phi_2^2 \chi^2 t}{2048 d x_\infty^2 x_{\max}^6 \sigma^2}}.
\end{align*}
\end{proposition}
\begin{proposition}\label{prop: beta 2 hat convergence 2}
For any $t\in [T]$,
$$\mathbb{P}\prns{\max_{x\in \mathcal{X}_1} |\hat{Q}_{t,2}(x,a)-Q_2(x,a)| > \chi, \mathcal{M}_{t}}\le 2d\exp\prns{-\frac{\phi_2^2 \chi^2 t}{32 d x_{\max}^2 x_{\infty}^2 \sigma^2}}.$$
\end{proposition}
}

What remains is to prove that $\mathcal{M}_t^C$ happens with low probability. 
Consider the following subsets of $\mathcal{S}_{a,m}$:
\begin{align}
 &\mathcal{S}_{a,1}'(t) = \braces{j\in [t]: j\not\in \cup_{a_1, a_2 \in[2]} \mathcal{T}_{(a_1, a_2)}, \bh_{j,1}\in U_{a,1}, \mathcal{G}_{j-1,1}}, \label{eq: 'sa1t} \\
 &\mathcal{S}'_{a,2}(t) =  \bigcup_{i\in[2]}\braces{ j \in [t]: j\not\in \cup_{a_1, a_2 \in[2]} \mathcal{T}_{(a_1, a_2)},   \bh_{j,1}\in U_{i,1} , \bh_{j,2}(i) \in U_{a,2}, \mathcal{G}_{j-1}} . \label{eq: 'sa2t}
\end{align}
\cref{lemma: s' iid} shows that $\braces{\bh_{j,m}: j\in \mathcal{S}_{a,m}'(t)}$ are independent samples.
We can then prove that under \cref{assumption: diversity}, $\abs{\mathcal{S}_{a,m}'(t)}< p_m t/4$ happens with low probability.
Moreover, we can apply the matrix Chernoff inequality to show that $\lambda_{\min}\prns{\sum_{s\in \mathcal{S}'_{a,m}(t)} \bh_{s,m} \bh_{s,m}^\intercal} $ concentrates around the smallest eigenvalue of the mean of $\sum_{s\in \mathcal{S}'_{a,m}(t)} \bh_{s,m} \bh_{s,m}^T$, which grows linearly when $\abs{\mathcal{S}_{a,m}'(t)}= \Omega(t)$.
Using the trivial relationship that $\lambda_{\min}\prns{\hat{\Sigma}_{a,m}(t)}\ge \lambda_{\min}\prns{\sum_{s\in \mathcal{S}'_{a,m}(t)} \bh_{s,m} \bh_{s,m}^\intercal}$, since all summand matrices are positive semidefinite, we can get the following bound on $\mathbb{P}(\mathcal{M}_t^C)$:
\begin{proposition} \label{lemma: prob on mt c}
When $q\geq \tilde{C}_1 \vee \tilde{C}_2$, $t\ge 256 q^2$ and $t\in\prns{\cap_{a_1, a_2 \in[2]} \mathcal{T}_{(a_1, a_2)}^C} \cup \{2^j q\}_{j\ge 5} $ (\ie, when we do not force-pull, or when we are at the last step of a force-pull stage), 
$$ \mathbb{P}\prns{\mathcal{M}_{t}^C}
    \le  \sum_{m\in[2]} 2d \exp\prns{-\frac{\phi_m t}{48 x_{\max}^2}} + \frac{280}{t^2}.$$
\end{proposition}

\subsection{Regret Upper Bound} 
Finally, we can compute the upper bound on our algorithm's regret.
Define $\mathcal{T}_0 = \braces{t: t\le 256 q^2} \cup \mathcal{T}_{1,1} \cup \mathcal{T}_{2,1}$.
We then decompose the regret of our algorithm into four exhaustive cases:
\begin{enumerate}
    \item $R^{(1)}$: when we initialize or explore ($t \in \mathcal{T}_0$);
    \item $R^{(2)}$: when $t\not\in \mathcal{T}_0$, and $\mathcal{G}_{t-1} \cap \mathcal{M}_{t-1} $ does not hold;
    \item $R^{(3)}$: when $t\not\in \mathcal{T}_0$, $\mathcal{G}_{t-1} \cap \mathcal{M}_{t-1} $ holds, and we pull the suboptimal arm in the first stage;
    \item $R^{(4)}$: when $t\not\in \mathcal{T}_0$, $\mathcal{G}_{t-1} \cap \mathcal{M}_{t-1} $ holds, and we pull the suboptimal arm in the second stage.
\end{enumerate}
Our regret is the sum of these four terms. To analyze our regret, we next show how control each of the terms, $R^{(1)}$ to $R^{(4)}$.

First, note that the per-step regret is always bounded by $6b_{\max} x_{\max}$. Because we have at most $3 q\log t$ forced pulls on each arm up to time $t$ when $t\ge 64 q^2$ (\cref{lemma: forced pull numbers}), it is easy to see that $R^{(1)} \le 8 q b_{\max} x_{\max} (3\log T + 128q)$.

Next, \cref{prop: convergence of Tilde Q,lemma: prob on mt c} guarantee that $\mathbb{P}\prns{\mathcal{G}_{t}^C\cup \mathcal{M}_{t}^C}$ is of order $O(1/t^2)$, so summing up over $t\not\in \mathcal{T}_0$ we get $R^{(2)} \le 24 b_{\max} x_{\max}\prns{49+ 16dx_{\max}^2/\phi_1 + 16d x_{\max}^2/\phi_2}$.

Thirdly, when $\mathcal{G}_{t-1} \cap \mathcal{M}_{t-1} $ holds, we never make a mistake when we use only the $\Tilde{Q}$ estimators to make a decision, so regret comes only when the $\Tilde{Q}$ estimators are close and we use $\hat{Q}$ estimators to make decisions. When $\bh_{t,1}$ falls into the region where $|Q_1(\bh_{t,1},1)-Q_1(\bh_{t,1},2)|=O\prns{t^{-1/2}}$, our algorithm may pull the suboptimal arm. However, such event happens with low probability $O(t^{-\alpha_1/2})$ by \cref{assumption: margin}, and the regret we incur in such an event is of order $O(t^{-1/2})$, so the total regret caused by such event is controllable. On the other hand, when $\bh_{t,1}$ falls into the region where $|Q_1(\bh_{t,1},1)-Q_1(\bh_{t,1},2)|=\Omega\prns{t^{-1/2}}$, by \cref{prop: convergence of hat Q 1} we very rarely make a mistake and pull the suboptimal arm at time $t$. Therefore, using a peeling argument, we obtain $R^{(3)}\le2^{\alpha_1+2}\gamma_1  C_1^{-\frac{\alpha_1+1}{2}} \prns{(M+2)^{\alpha_1+2}+ 2^{\alpha_1+6}}f(\alpha_1)$.

The analysis of $R^{(4)}$ is almost analogous to the analysis of $R^{(3)}$, and we get $R^{(4)}\le 2^{\alpha_2+3}\gamma_2  C_2^{-\frac{\alpha_2+1}{2}} \prns{(M+2)^{\alpha_2+2}+ 2^{\alpha_2+4}}f(\alpha_2)$. Adding all four terms together we get the result in \cref{thm: regret upper bound}.

\section{Omitted Proofs}\label{section: proof}
In this section, any $a$ denotes one of the arms and can take value in $[2]$.

\subsection{Proofs in \cref{section: formulation}}
{
\begin{myproof}[Proof of \cref{prop: expected regret}.]
On one hand,
\begin{align*}
   \expect\bracks{Y_t(A^*_{t,1}, A^*_{t,2})}
  = & \expect\bracks{Q_2(\bh^*_{t,2}, A^*_{t,2})} \\
  = & \expect\bracks{\max_{a_2} Q_2(\bh^*_{t,2}, a_2)}\\
  = & \expect\bracks{\expect\bracks{\max_{a_2} Q_2(\bh^*_{t,2}, a_2)\mid \bx_{t,1}, A^*_{t,1}}}\\
  = & \expect\bracks{Q_1(\bx_{t,1}, A^*_{t,1})}.
\end{align*}
On the other hand,
\begin{align*}
   \expect\bracks{Y_t(A_{t,1}, A_{t,2}) }
   = &\expect\bracks{Q_2(\bh_{t,2}, A_{t,2}))}\\
   = & \expect\bracks{\max_{a_2}Q_2(\bh_{t,2}, a_2)} - \expect\bracks{\max_{a_2}Q_2(\bh_{t,2}, a_2)} + \expect\bracks{Q_2(\bh_{t,2}, A_{t,2})}\\
   = & \expect\bracks{Q_1(\bx_{t,1}, A_{t,1})} - \expect\bracks{\max_{a_2}Q_2(\bh_{t,2}, a_2)} + \expect\bracks{Q_2(\bh_{t,2}, A_{t,2})}.
\end{align*}
Taking a difference between these two we get the desired result.
\end{myproof}

\begin{myproof}[Proof of \cref{lemma: alpha is one}]
We will give a detailed proof for $\alpha_1 = 1$. The arguments for $\alpha_2 = 1$ are similar.

For simplicity suppose $\bb_{1,1} \neq \bb_{2,1}$; otherwise pulling either arm is optimal in stage $1$.
Letting $V_d(R)$ be the volume of the $R$-radius $d$-ball, we have
\begin{align*}
    \pr\prns{0<\abs{Q_1\prns{\bx_{t,1}, 1} - Q_1\prns{\bx_{t,1}, 2}} \le \kappa} \le & 2\pr\prns{0<\prns{\bb_{1,1} - \bb_{2,1}}^\intercal\bp_1\prns{\bx_{t,1}} \le \kappa} \\
    \le & 2\mu_{\max} \int_0^{\kappa/\norm{\bb_{1,1} - \bb_{2,1}}} V_{d-1} \prns{\prns{x_{\max}^2 - u^2}_+^{1/2}}d u\\
    \le & 12\mu_{\max} x_{\max}^d \kappa / \norm{\bb_{1,1} - \bb_{2,1}}.
\end{align*}
\end{myproof}

}
\subsection{Proofs for Regret Upper Bound}

\subsubsection{Proofs in \cref{section: two tails}}

{

\begin{myproof}[Proof of \cref{prop: ols}.]
It is easy to check that $\hat{\bb}-\bb = \hat{\Sigma}^{-1}\prns{\sum_{i=1}^n g(\bh_i) \epsilon_i + \sum_{i=1}^n g(\bh_i) \delta_i } $.
When the event $\lambda_{\min}\prns{\hat{\Sigma}}\ge \phi$ holds,
\begin{align*}
   \norm{\hat{\bb}-\bb} & = \norm{\hat{\Sigma}^{-1}\prns{\sum_{i=1}^n g(\bh_i) \epsilon_i + \sum_{i=1}^n g(\bh_i) \delta_i } } \\
    & \le \norm{\hat{\Sigma}^{-1}}\cdot\prns{\norm{\sum_{i=1}^n g(\bh_i) \epsilon_i } + \norm{\sum_{i=1}^n g(\bh_i) \delta_i }}\\
    & \le \frac{1}{ \phi}\norm{\sum_{i=1}^n g(\bh_i) \epsilon_i } + \frac{1}{\phi}\norm{\sum_{i=1}^n g(\bh_i) \delta_i }\\
    & \le \frac{1}{ \phi}\norm{\sum_{i=1}^n g(\bh_i) \epsilon_i } + \frac{x_{\max}}{\phi}\sum_{i=1}^n\abs{ \delta_i }.
\end{align*}
Therefore, for any $\chi>0$,
$$\pr\prns{\norm{\hat{\bb}-\bb}\ge \chi, \lambda_{\min}\prns{\hat{\Sigma}}\ge \phi}  \le \pr\prns{\norm{\sum_{i=1}^n g(\bh_i) \epsilon_i } \ge \frac{\chi\phi}{2}} + \pr\prns{\sum_{i=1}^n\abs{ \delta_i }\ge \frac{\chi \phi}{2x_{\max}}}.$$
Note that
$$
\pr\prns{\norm{\sum_{i=1}^n g(\bh_i) \epsilon_i } \ge \frac{\chi\phi}{2}} 
\le 
\sum_{j=1}^d\pr\prns{\abs{\sum_{i=1}^n g_j(\bh_i) \epsilon_i } \ge \frac{\chi\phi}{2\sqrt{d}}} .
$$
Because each $g_j(\bh_i) \epsilon_i$ is $x_{\infty}\sigma$-sub-Gaussian and $\{g_j(\bh_i) \epsilon_i\}_{i=1}^n$ is a martingale difference sequence adapted to the filtration $\{\mathcal{F}_i\}_{i=0}^n$, by \citet[theorem 2.19]{wainwright2019high},
$$
\pr\prns{\abs{\sum_{i=1}^n g_j(\bh_i) \epsilon_i } \ge \frac{\chi\phi}{2\sqrt{d}}} \le 2\exp\prns{-\frac{\chi^2 \phi^2}{8n d x_{\infty}^2 \sigma^2}}.
$$
Combining all above pieces we complete the proof.

\end{myproof}
}

\subsubsection{Proofs in \cref{section: tilde estimators}}
\label{section: proof tilde}
First, we prove a supporting lemma on that bounds the number of forced pulls. Its proof is modified from lemma EC.8 in \citet{bastani2020online}.
{
\begin{lemma}\label{lemma: forced pull numbers}
If $t\ge 64 q^2$, then $q\log t \le |\mathcal{T}_{a,m}(t)| \le 3q\log t$.
\end{lemma}
\begin{myproof}[Proof of \cref{lemma: forced pull numbers}]
Define the $n$-th consecutive round of forced pulls as
$$L_n = \braces{\prns{2^{n+2}-4}q + 1, \dots, 2^{n+2}q }, \quad n\ge 0.$$
By construction, each arm $a$ is sampled $2q $ times in the $m$-th stage during $L_n$. 
Let $n_t = \sup\braces{j\in \mathbb{N}: t\ge 2^{j+2} q}$, and we have $2n_t q \le \abs{\mathcal{T}_{a,m}(t)} \le 2(n_t+1)q$.

To obtain the lower bound, note that $t\le 2^{n_t+3} q$, which implies $n_t\ge \log(t/8q)$. For $t\ge 64q^2$,
$$\abs{\mathcal{T}_{a,m}(t)} \ge 2n_t q \ge 2q \log\prns{t/8q} \ge 2q\prns{\log t - \log (8q) } \ge q\log t.$$

To obtain the upper bound, note that $t\ge 2^{n_t+2} q$, which implies $n_t\le \log_2(t/4q)$.
$$\abs{\mathcal{T}_{a,m}(t)} \le 2\prns{n_t+1} q \le 2\log_2(t/2q) q \le 3q\log t.$$
\end{myproof}
}

\paragraph{I. Positive Definiteness of Design Matrices.}

\begin{lemma} \label{lemma: positive defnite 1}
For any $t\in [T]$,
$$\mathbb{P} \prns{\lambda_{\min}\prns{\Tilde{\Sigma}_{a,1}(t)} \le \frac{\phi_1}{2}\abs{\mathcal{T}_{a,1}(t)} } \le  d\exp\prns{-\frac{\abs{\mathcal{T}_{a,1}(t)}\phi_1}{8  x_{\max}^2}}.$$
\end{lemma}
\begin{myproof}[Proof of \cref{lemma: positive defnite 1}]
First we have
\begin{align*}
     \lambda_{\max} \prns{\bp_1\prns{\bh_{j,1}} \bp_1^\intercal\prns{\bh_{j,1}}} =   \max_{\norm{u}=1}  u^T\bp_1\prns{\bh_{j,1}} \bp_1^\intercal\prns{\bh_{j,1}} u \le  x_{\max}^2.
\end{align*}
Moreover,
\begin{align*}
   \lambda_{\min}\prns{\expect \bracks{\Tilde{\Sigma}_{a,1}(t)} }  & = \lambda_{\min}\prns{\sum_{j\in \mathcal{T}_{a,1}(t)}\expect \bracks{\bp_1\prns{\bh_{j,1}} \bp_1^\intercal\prns{\bh_{j,1}}} }\\
   & \ge \lambda_{\min}\prns{\sum_{j\in \mathcal{T}_{a,1}(t)}\expect \bracks{\bp_1\prns{\bh_{j,1}} \bp_1^\intercal\prns{\bh_{j,1}}} \ind \braces{\bh_{j,1} \in U_{a,1}} }\\
   & \ge \sum_{j\in \mathcal{T}_{a,1}(t)} \lambda_{\min}\prns{\expect \bracks{\bp_1\prns{\bh_{j,1}} \bp_1^\intercal\prns{\bh_{j,1}}} \ind \braces{\bh_{j,1} \in U_{a,1}} }
  \\
   & \ge \phi_1\abs{\mathcal{T}_{a,1}(t)}.
\end{align*}
Because $ \braces{\bp_1\prns{\bh_{j,1}} \bp_1^\intercal\prns{\bh_{j,1}}}_{j\in \mathcal{T}_{a,1}(t)}$ are independent, by Matrix Chernoff inequality \citep[\cf][Theorem 5.1.1]{tropp2015introduction},
$$\mathbb{P} \prns{\lambda_{\min}\prns{\Tilde{\Sigma}_{a,1}(t)} \le \frac{\phi_1}{2}\abs{\mathcal{T}_{a,1}(t)} } \le d\exp\prns{-\frac{\abs{\mathcal{T}_{a,1}(t)}\phi_1}{8  x_{\max}^2}} .$$
\end{myproof}

\begin{lemma}\label{lemma: positive defnite 2}
For any $t\in [T]$,
$$\mathbb{P} \prns{\lambda_{\min}\prns{\Tilde{\Sigma}_{a,2}(t)} \le \frac{\phi_2}{4}\abs{\mathcal{T}_{a,2}(t)} } \le  d\exp\prns{-\frac{\abs{\mathcal{T}_{a,2}(t)}\phi_2}{{16}  x_{\max}^2}}.$$
\end{lemma}
\begin{myproof}[Proof of \cref{lemma: positive defnite 2}]
By the same argument as in \cref{lemma: positive defnite 1}, we know $\lambda_{\max}\prns{\bp_2\prns{\bh_{j,2}} \bp_2^\intercal\prns{\bh_{j,2}}}\le  x_{\max}^2$. {
Note that
\begin{align*}
   \Tilde{\Sigma}_{a,2}(t)
   = &\sum_{j\in \mathcal{T}_{a,2}(t)} \bp_2\prns{\bh_{j,2}} \bp_2^\intercal\prns{\bh_{j,2}} \\
    = & \sum_{j\in \mathcal{T}_{a,2}(t)\cap \mathcal{T}_{1,1}(t)} \bp_2\prns{\bh_{j,2}(1)} \bp_2^\intercal\prns{\bh_{j,2}(1)} + \bp_2\prns{\bh_{j+2q,2}(2)} \bp_2^\intercal\prns{\bh_{j+2q,2}(2)},
\end{align*}
where the last equality follows from our forced-pull schedule.
Since
\begin{align*}
   & \lambda_{\min}\prns{ \sum_{j\in \mathcal{T}_{a,2}(t)\cap \mathcal{T}_{1,1}(t)} \expect\bracks{\bp_2\prns{\bh_{j,2}(1)} \bp_2^\intercal\prns{\bh_{j,2}(1)} + \bp_2\prns{\bh_{j+2q,2}(2)} \bp_2^\intercal\prns{\bh_{j+2q,2}(2)}} } \\
   = & \lambda_{\min}\prns{ \sum_{j\in \mathcal{T}_{a,2}(t)\cap \mathcal{T}_{1,1}(t)} \sum_{a_1\in[2]} \expect\bracks{\bp_2\prns{\bh_{j,2}(a_1)} \bp_2^\intercal\prns{\bh_{j,2}(a_1)}} }\\
   \ge & \sum_{j\in \mathcal{T}_{a,2}(t)\cap \mathcal{T}_{1,1}(t)} \lambda_{\min}\prns{  \sum_{a_1\in[2]} \expect\bracks{\bp_2\prns{\bh_{j,2}(a_1)} \bp_2^\intercal\prns{\bh_{j,2}(a_1)}} }\\
   \ge & \frac{\phi_2}{2} \abs{\mathcal{T}_{a,2}(t)},
\end{align*}
}
and each term in the sum is independent, by Matrix Chernoff inequality,
\begin{align*}
\mathbb{P} \prns{\lambda_{\min}\prns{\Tilde{\Sigma}_{a,2}(t)} \le \frac{\phi_2}{4}\abs{\mathcal{T}_{a,2}(t)} } \le  d\exp\prns{-\frac{\abs{\mathcal{T}_{a,2}(t)}\phi_2}{16  x_{\max}^2}}.
\end{align*}
\end{myproof}

\paragraph{II. Concentration of $\Tilde{Q}_{t,1}$ and $\Tilde{Q}_{t,2}$.}
{
\begin{myproof}[Proof of \cref{prop: convergence of Tilde Q}.]
First of all, for any $\chi>0$,
\begin{align}
    & \pr\prns{\norm{\tilde{\bb}_{a_2, 2} (t)-\bb_{a_2,2}} \ge \chi } \notag\\
    \le & \pr\prns{ \norm{\tilde{\bb}_{a_2,2}(t) - \bb_{a_2,2}} \ge \chi, \lambda_{\min}\prns{\Tilde{\Sigma}_{a_2,2}(t)} \ge \frac{\phi_2}{4}\abs{\mathcal{T}_{a_2,2}(t)}} + d\exp\prns{-\frac{\abs{\mathcal{T}_{a_2,2}(t)}\phi_2}{16  x_{\max}^2}} \notag\\
    \le & 2d\exp\prns{-\frac{\chi^2\phi_2^2 \abs{\mathcal{T}_{a_2,2}(t)}}{128dx_{\infty}^2 \sigma^2}} +  d\exp\prns{-\frac{\abs{\mathcal{T}_{a_2,2}(t)}\phi_2}{16  x_{\max}^2}} \label{eq: beta 2},
\end{align}
where the first inequality follows from \cref{lemma: positive defnite 2}, and the second inequality follows from \cref{prop: ols}.
Statement \ref{eq: gt2} is a direct consequence of this result:
\begin{align*}
     & \pr\prns{\max_{a_2\in[2]} \sup_{\textbf{h} \in \mathcal{H}_2} \abs{\tilde{Q}_{t,2}(\textbf{h},a_2) - Q_1(\textbf{h}, a_2)} > \frac{\Delta_2}{4}} \\
    \le  & \sum_{a_2\in [2]}\pr\prns{ \norm{\tilde{\bb}_{a_2,2}(t) - \bb_{a_2,2}} > \frac{\Delta_2}{4 x_{\max}}} \\
    \le & 4d\exp\prns{-\frac{\Delta_2^2\phi_2^2 \abs{\mathcal{T}_{a_2,2}(t)}}{2048d x_{\max}^2 x_{\infty}^2 \sigma^2}} +  2d\exp\prns{-\frac{\abs{\mathcal{T}_{a_2,2}(t)}\phi_2}{16  x_{\max}^2}} \\
    \le & \frac{6}{t^2},
\end{align*}
where the last inequality follows from \cref{lemma: forced pull numbers} and $q\ge \tilde{C}_1$.

We now prove Statement \ref{eq: gt1}. Note that
\begin{align*}
    & \pr\prns{\max_{a_1\in[2]}\sup_{\textbf{h} \in \mathcal{H}_1} \abs{\tilde{Q}_{t,1}(\textbf{h},a_1) - Q_1(\textbf{h}, a_1)} > \frac{\Delta_1}{4}} \\
    \le  & \sum_{a_1\in [2]}\pr\prns{ \norm{\tilde{\bb}_{a_1,1}(t) - \bb_{a_1,1}} > \frac{\Delta_1}{4 x_{\max}}} \\
    \le & \sum_{a_1\in [2]} \pr\prns{ \norm{\tilde{\bb}_{a_1,1}(t) - \bb_{a_1,1}} > \frac{\Delta_1}{4 x_{\max}}, \lambda_{\min}\prns{\Tilde{\Sigma}_{a_1,1}(t)} > \frac{\phi_1}{2}\abs{\mathcal{T}_{a_1,1}(t)}} + d\exp\prns{-\frac{\abs{\mathcal{T}_{a_1,1}(t)}\phi_1}{8  x_{\max}^2}},
\end{align*}
where the last inequality follows from a union bound and \cref{lemma: positive defnite 1}.
For any $j\in \mathcal{T}_{a_1,1}(t)$,
\begin{align*}
     \tilde{Y}_j = & \max_{a\in[2]} \tilde{\bb}_{a,2}^\intercal(t)\bp_2\prns{\bh_{j,2}} \\
    = & \bb_{a_1,1}^\intercal\bp\prns{\bx_{t,1}} + \bracks{ \max_{a\in[2]} \bb_{a,2}^\intercal\bp_2\prns{\bh_{j,2}}  - \bb_{a_1,1}^\intercal\bp\prns{\bx_{t,1}}} +\bracks{\max_{a\in[2]} \tilde{\bb}_{a,2}^\intercal(t)\bp_2\prns{\bh_{j,2}} - \max_{a\in[2]} \bb_{a,2}^\intercal\bp_2\prns{\bh_{j,2}} } \\
    = & \bb_{a_1,1}^\intercal\bp\prns{\bx_{t,1}} + \eta_{j,1}^{(a_1)} +\bracks{\max_{a\in[2]} \tilde{\bb}_{a,2}^\intercal(t)\bp_2\prns{\bh_{j,2}} - \max_{a\in[2]} \bb_{a,2}^\intercal\bp_2\prns{\bh_{j,2}} }.
\end{align*}
By \cref{prop: ols}, 
\begin{align*}
     & \pr\prns{ \norm{\tilde{\bb}_{a_1,1}(t) - \bb_{a_1,1}} > \frac{\Delta_1}{4 x_{\max}}, \lambda_{\min}\prns{\Tilde{\Sigma}_{a_1,1}(t)} > \frac{\phi_1}{2}\abs{\mathcal{T}_{a_1,1}(t)}} \\
     \le & 2d\exp\prns{-\frac{\Delta_1^2 \phi_1^2 \abs{\mathcal{T}_{a_1,1}(t)}}{512 d x_{\max}^2 x_{\infty}^2 \sigma^2}}  + \pr\prns{\sum_{j \in \mathcal{T}_{a_1, 1}(t)}\abs{\max_{a\in[2]} \tilde{\bb}_{a,2}^\intercal(t)\bp_2\prns{\bh_{j,2}} - \max_{a\in[2]} \bb_{a,2}^\intercal\bp_2\prns{\bh_{j,2}} } \ge \frac{\Delta_1 \phi_1\abs{\mathcal{T}_{a_1, 1}(t)} }{16 x_{\max}^2}  }\\
     \le & 2d\exp\prns{-\frac{\Delta_1^2 \phi_1^2 \abs{\mathcal{T}_{a_1,1}(t)}}{512 d x_{\max}^2 x_{\infty}^2 \sigma^2}} + \sum_{a\in [2]}\pr\prns{\norm{\tilde{\bb}_{a, 2} -\bb_{a,2}} \ge \frac{\Delta_1 \phi_1}{16 x_{\max}^3} }\\
     \le & 2d\exp\prns{-\frac{\Delta_1^2 \phi_1^2 \abs{\mathcal{T}_{a_1,1}(t)}}{512 d x_{\max}^2 x_{\infty}^2 \sigma^2}} + \sum_{a\in [2]} 2d\exp\prns{-\frac{\Delta_1^2 \phi_1^2\phi_2^2 \abs{\mathcal{T}_{a,2}(t)}}{32768 d x_{\max}^6  x_{\infty}^2 \sigma^2}} +  d\exp\prns{-\frac{\abs{\mathcal{T}_{a,2}(t)}\phi_2}{16  x_{\max}^2}}\\
     \le & \frac{8}{t^2},
\end{align*}
where the last inequality follows from \cref{lemma: forced pull numbers}, $q\ge \tilde{C}_1 \vee \tilde{C}_2$ .
\end{myproof}
}

\subsubsection{Proofs in \cref{section: hat estimators}}
\label{section: proof hat}

\paragraph{I. Characterization of $\mathcal{S}_{a,m}'(t)$.}

\begin{lemma}\label{lemma: s inclusion}
For any $t\in [T]$, $a, m\in [2]$,
$$\mathcal{S}_{a,m}'(t)\subseteq \mathcal{S}_{a,m}(t).$$
\end{lemma}
\begin{myproof}[Proof of \cref{lemma: s inclusion}]
If we have $\max_{a\in [2]}\sup_{\textbf{h}\in \mathcal{H}_1} |\Tilde{Q}_{j-1,1}(\textbf{h},a)-Q_1(\textbf{h},a)| \le \Delta_1/4$ (\ie, $\mathcal{G}_{j-1,1}$) and $Q_1(\bh_{j,1},a)> Q_1(\bh_{j,1}, 3-a) + \Delta_1$ (\ie, $\bh_{j,1}\in U_{a,1}$), we know
\begin{align*}
  & \Tilde{Q}_{j-1,1}(\bh_{j,1},a)-\Tilde{Q}_{j-1,1}(\bh_{j,1}, 3-a)\\
  > & \Tilde{Q}_{j-1,1}(\bh_{j,1},a) -Q_1(\bh_{j,1},a) + Q_1(\bh_{j,1}, 3-a) -\Tilde{Q}_{j-1,1}(\bh_{j,1}, 3-a) +\Delta_1\\
  > & \frac{\Delta_1}{2}.
\end{align*}
By our algorithm, whenever $j \not\in\cup_{a_1, a_2 \in[2]} \mathcal{T}_{(a_1, a_2)}$ and $\Tilde{Q}_{j-1,1}(\bh_{j,1},a)-\Tilde{Q}_{j-1,1}(\bh_{j,1}, 3-a)>\Delta_1/2$, we pull arm $a$.
Therefore, $\mathcal{S}_{a,1}'(t) \subseteq\mathcal{S}_{a,1}(t)$.

Define
$$\mathcal{S}'_{(i,a)}(t) =  \braces{ j \in [t]: j\not\in \cup_{a_1, a_2 \in[2]} \mathcal{T}_{(a_1, a_2)},   \bh_{j,1}\in U_{i,1} , \bh_{j,2}(i) \in U_{a,2}, \mathcal{G}_{j-1}},$$
so we have $\mathcal{S}'_{a,2}(t) = \cup_{i\in[2]}\mathcal{S}'_{(i,a)}(t)$.
Because $\mathcal{S}_{(i,a)}'(t)\subseteq\mathcal{S}_{i,1}'(t)\subseteq\mathcal{S}_{i,1}(t)$, for any $j\in \mathcal{S}_{(i,a)}'(t)$, we have $\bh_{j,2} =\bh_{j,2}(i)$. Moreover, $\mathcal{G}_{j-1,2}$ and $\bh_{j,2}\in U_{a,2}$ implies $\Tilde{Q}_{j-1,2}(\bh_{j,2},a)-\Tilde{Q}_{j-1,2}(\bh_{j,2}, 3-a)>\Delta_2/2$, and by our algorithm $A_{j,2} = a$. Thus $\mathcal{S}'_{(i,a)}(t) \subseteq \mathcal{S}_{a,2}(t)$, so $\mathcal{S}'_{a,2}(t) \subseteq \mathcal{S}_{a,2}(t)$.
\end{myproof}

\begin{lemma} \label{lemma: s' iid}
$\{\bh_{j,1}: j\in \mathcal{S}_{a,1}'(t)\}$ are i.i.d. and $\{\bh_{j,2}: j\in \mathcal{S}_{a,2}'(t)\}$ are independent samples.
\end{lemma}
\begin{myproof}[Proof of \cref{lemma: s' iid}]
Note that $j\not\in \cup_{a_1, a_2 \in[2]} \mathcal{T}_{(a_1, a_2)}$ is deterministic. Because $\mathcal{G}_{j-1,1}$ and $\mathcal{G}_{j-1}$ only depend on samples in $\mathcal{T}_{1,1}(j-1) \cup \mathcal{T}_{2,1}(j-1)$, they are independent of $\bh_{j,1}$, $\bh_{j,2}(1)$ and $\bh_{j,2}(2)$. {Therefore, $\{\bh_{j,1}: j\not\in \cup_{a_1, a_2 \in[2]} \mathcal{T}_{(a_1, a_2)}, \mathcal{G}_{j-1,1} \}$ are i.i.d. samples from $\mathcal{P}_{\bx_1}$, and $\{(\bh_{j,1}, \bh_{j,2}(1),\bh_{j,2}(2)): j\not\in \cup_{a_1, a_2 \in[2]} \mathcal{T}_{(a_1, a_2)}, \mathcal{G}_{j-1} \}$ are i.i.d. samples from $\mathcal{P}_{(\bx_1 ,\bh_2(1), \bh_2(2))}$. Besides, whether $\bh_{j,1}$ falls in $U_{a,1}$ and whether $\cup_{i\in[2]} \braces{\bh_{j,1}\in U_{i,1} , \bh_{j,2}(i) \in U_{a,2}}$ holds are simply rejection sampling}, so the results follow.
\end{myproof}

\paragraph{II. Bounding $\mathcal{M}_t^C$}
\begin{myproof}[Proof of \cref{lemma: prob on mt c}]
Note that for any $n\in \{0,1,2,\dots\}$, we do not perform any forced-sampling in {$\bracks{2^{n+1}q+1, \prns{2^{n+2}-4} q}$}. For any $t\in\prns{\cap_{a_1, a_2 \in[2]} \mathcal{T}_{(a_1, a_2)}^C} \cup \{2^j q\}_{j\ge 5} $ , define
\begin{equation} \label{eq: nt}
    n_t = \sup\braces{j\in \mathbb{N}_+: t\ge 2^{j+2}q},
\end{equation}
\begin{equation} \label{eq: vt}
    V_t = \bracks{2^{n_t+1}q+1, \prns{2^{n_t+2}-4} q}\cup \bracks{2^{n_t+2}q+1,t} .
\end{equation}
When $t\ge 32q$,
\begin{align*}
    |V_t| & = t - 2^{n_t+1}q-4q  \ge \frac{t}{2}-4q \ge \frac{3}{8}t.
\end{align*}

Let
$$N_{a,m}(t)\triangleq \mathcal{S}'_{a,m}(t) \cap V_t \subseteq \mathcal{S}'_{a,m}(t).$$
We have
{
\begin{align*}
    \ind\braces{s\in N_{a,1}(t)} = & \ind\braces{\mathcal{G}_{s-1,1}} \ind\braces{\bh_{s,1}\in U_{a,1}}\ind\braces{s \in V_t}\\
    \ge & \ind\braces{\mathcal{G}_{2^{n_t+1}q}} \ind\braces{\mathcal{G}_{2^{n_t+2}q}} \ind\braces{\bh_{s,1}\in U_{a,1}}\ind\braces{s \in V_t},
\end{align*}
where the inequality follows from the fact that when $s\in \bracks{2^{n_t+1}q+1, \prns{2^{n_t+2}-4} q}$, $\mathcal{G}_{s-1}=\mathcal{G}_{2^{n_t+1}q}$ and when $s\in \bracks{2^{n_t+2}q+1,t}$, $\mathcal{G}_{s-1}=\mathcal{G}_{2^{n_t+2}q}$.}

Note that
\begin{align*}
    \lambda_{\min}\prns{\hat{\Sigma}_{a,m}(t)}   = & \lambda_{\min}\prns{\sum_{s\in \mathcal{S}_{a,m}(t)} \bp_m\prns{\bh_{s,m} } \bp_m^\intercal\prns{\bh_{s,m} }}\\  \ge & \lambda_{\min}\prns{\sum_{s\in N_{a,m}(t)} \bp_m\prns{\bh_{s,m} } \bp_m^\intercal\prns{\bh_{s,m} }}.
\end{align*}
Moreover, we have
\begin{align*}
  & \lambda_{\min}\prns{\expect\bracks{\sum_{s\in N_{a,1}(t)} \bp_1\prns{\bh_{s,1} } \bp_1^\intercal\prns{\bh_{s,1} }   \mid \mathcal{G}_{2^{n_t+1}q} \cap \mathcal{G}_{2^{n_t+2}q}}} \\
  \ge &  \lambda_{\min}\prns{\expect\bracks{\sum_{s\in V_t} \bp_1\prns{\bh_{s,1} } \bp_1^\intercal\prns{\bh_{s,1} } \ind\braces{\bh_{s,1}\in U_{a,1}}  \ind\braces{\mathcal{G}_{2^{n_t+1}q}} \ind\braces{\mathcal{G}_{2^{n_t+2}q}} \mid \mathcal{G}_{2^{n_t+1}q} \cap \mathcal{G}_{2^{n_t+2}q}}} \\
  = &  \lambda_{\min}\prns{\expect\bracks{\sum_{s\in V_t} \bp_1\prns{\bh_{s,1} } \bp_1^\intercal\prns{\bh_{s,1} } \ind\braces{\bh_{s,1}\in U_{a,1}} }} \\
   \ge & \frac{ 3\phi_1 t}{8},
\end{align*}
and, by similar arguments,
$$  \lambda_{\min}\prns{\expect\bracks{\sum_{s\in N_{a,2}(t)} \bp_2\prns{\bh_{s,2} } \bp_2^\intercal\prns{\bh_{s,2} }   \mid \mathcal{G}_{2^{n_t+1}q} \cap \mathcal{G}_{2^{n_t+2}q}}}  \ge \frac{3\phi_2 t}{8}.$$
By matrix Chernoff inequality \citep[\cf][Theorem 5.1.1]{tropp2015introduction},
\begin{align*}
    & \pr\prns{\lambda_{\min}\prns{\hat{\Sigma}_{a,m}(t)} \le \frac{\phi_m t}{4},  \mathcal{G}_{2^{n_t+1}q} \cap \mathcal{G}_{2^{n_t+2}q}}\\
    \le & \pr\prns{\lambda_{\min}\prns{\sum_{s\in N_{a,m}(t)} \bp_m\prns{\bh_{s,m} } \bp_m^\intercal\prns{\bh_{s,m} }} \le \frac{\phi_m t}{4},  \mathcal{G}_{2^{n_t+1}q} \cap \mathcal{G}_{2^{n_t+2}q}}\\
    \le & d \exp\prns{-\frac{\phi_m t}{48 x_{\max}^2}} .
\end{align*}

Finally, since $2^{n_t+1}q >t/4\ge 64q^2$, by \cref{prop: convergence of Tilde Q},
\begin{align*}
    \mathbb{P}\prns{\mathcal{G}_{2^{n_t+1}q}^C \cup \mathcal{G}_{2^{n_t+2}q}^C} \le \frac{{14}}{(t/4)^2}+\frac{{14}}{(t/2)^2}\le \frac{{280}}{t^2}.
\end{align*}

By a union bound we have
\begin{align*}
    \mathbb{P}\prns{\mathcal{M}_{t}^C}
    \le & 
   \sum_{m\in[2]} \sum_{a\in [2]}\mathbb{P}\prns{\lambda_{\min}\prns{\hat{\Sigma}_{a,m}(t)} < \frac{\phi_m t}{4}, \mathcal{G}_{2^{n_t+1}q} \cap \mathcal{G}_{2^{n_t+2}q}} 
    +   \mathbb{P}\prns{\mathcal{G}_{2^{n_t+1}q}^C \cup \mathcal{G}_{2^{n_t+2}q}^C}\\
    \le & \sum_{m\in[2]} 2d \exp\prns{-\frac{\phi_m t}{48 x_{\max}^2}} + \frac{280}{t^2}.
\end{align*}

\end{myproof}

\paragraph{III. Concentration of $\hat{Q}_{t,1}$ and $\hat{Q}_{t,2}$.}
{

\begin{myproof}[Proof of \cref{prop: convergence of hat Q 1,prop: beta 2 hat convergence 2}.]
First, following similar arguments as in the proof of \cref{prop: ols},
when the event $\lambda_{\min}\prns{\hat{\Sigma}_{a,2}(t)}\ge \phi $ holds,
\begin{align*}
    \norm{\hat{\bb}_{a,2}(t) - \bb_{a,2}}\le & \frac{1}{ \phi}\norm{\sum_{j\in \mathcal{S}_{a,2}(t)} \bp_2(\bh_{j,2})\eta_{j,2}^{(A_{j,1}, a)} } \\
    = & \frac{1}{ \phi}\norm{\sum_{j=1}^t \bp_2(\bh_{j,2})\eta_{j,2}^{(A_{j,1}, a)} \ind\braces{A_{j,2} = a} }.
\end{align*}
Let $\mathcal{F}_j$ be the $\sigma$-algebra generated by $\braces{\bh_{i,2}, A_{i,2}, Y_i}_{i\in[j]} \cup \bh_{j+1,2}$. If we denote the $k$-th entry of $\bp_2\prns{\bh_{j,2}}$ as $\bp_{2k}\prns{\bh_{j,2}}$, it is easy to see that $\braces{ \bp_{2k}(\bh_{j,2})\eta_{j,2}^{(A_{j,1}, a)} \ind\braces{A_{j,2} = a}}_{j=1}^T$ is a martingale difference sequence adapted to $\braces{\mathcal{F}_j}_{j=1}^T$.
Thus, for any $\chi'>0$,
\begin{align}
   & \pr\prns{\norm{\hat{\bb}_{a,2}(t) - \bb_{a,2}}> \chi', \lambda_{\min}\prns{\hat{\Sigma}_{a,2}(t)}\ge \phi} \notag\\
    \le & \sum_{k=1}^d \pr\prns{ \abs{\sum_{j=1}^t \bp_{2k}(\bh_{j,2})\eta_{j,2}^{(A_{j,1}, a)} \ind\braces{A_{j,2} = a} }>  \frac{\phi \chi'}{\sqrt{d}}} \notag\\
    \le & 2d \exp\prns{-\frac{\phi^2 \chi'^2}{2 t d x_\infty^2 \sigma^2}}. \label{eq: useful}
\end{align}
where the last inequality follows from \citet[theorem 2.19]{wainwright2019high}.

\cref{prop: beta 2 hat convergence 2} is a direct consequence of \cref{eq: useful}:
\begin{align*}
    & \pr\prns{\max_{\textbf{h}\in \mathcal{H}_2} \abs{\hat{Q}_{t,2}(\textbf{h},a)-Q_2(\textbf{h},a)}> \chi, \mathcal{M}_{t}}\\
    \le & \pr\prns{\norm{\hat{\bb}_{a,2}(t) - \bb_{a,2}}> \frac{\chi}{x_{\max}}, \lambda_{\min}\prns{\hat{\Sigma}_{a,2}(t)}\ge \frac{\phi_2 t}{4}} \\
    \le & 2d \exp\prns{-\frac{\phi_2^2  \chi^2 t}{32  d x_\infty^2 x_{\max}^2\sigma^2}}.
\end{align*}

Now we prove \cref{prop: convergence of hat Q 1}. Note that
\begin{align*}
    \pr\prns{\max_{\textbf{h}\in \mathcal{H}_1} \abs{\hat{Q}_{t,1}(\textbf{h},a)-Q_1(\textbf{h},a)}> \chi, \mathcal{M}_{t}} \le \pr\prns{\norm{\hat{\bb}_{a,1}(t) - \bb_{a,1}}> \frac{\chi}{x_{\max}}, \mathcal{M}_{t}}.
\end{align*}
Recall that $\hat{\bb}_{a, 1}(t) = \hat{\Sigma}_{a,1}^{-1}(t) \prns{\sum_{j\in \mathcal{S}_{a,1}(t)} \bp_1(\bh_{j,1})\hat{Y}_j  } $, where
\begin{align*}
    \hat{Y}_j = &\max_{a_2\in[2]} \hat{\bb}_{a_2, 2}^\intercal(t) \bp_2\prns{\bh_{j,2}} \\
    = & \bb_{a,1}^\intercal\bp\prns{\bh_{t,1}} + \bracks{ \max_{a_2\in[2]} \bb_{a_2,2}^\intercal\bp_2\prns{\bh_{j,2}}  - \bb_{a,1}^\intercal\bp\prns{\bh_{t,1}}} +\bracks{\max_{a_2\in[2]} \hat{\bb}_{a_2,2}^\intercal(t)\bp_2\prns{\bh_{j,2}} - \max_{a_2\in[2]} \bb_{a_2,2}^\intercal\bp_2\prns{\bh_{j,2}} } \\
    = & \bb_{a,1}^\intercal\bp\prns{\bh_{t,1}} + \eta_{j,1}^{(a)} +\bracks{\max_{a_2\in[2]} \hat{\bb}_{a_2,2}^\intercal(t)\bp_2\prns{\bh_{j,2}} - \max_{a_2\in[2]} \bb_{a_2,2}^\intercal\bp_2\prns{\bh_{j,2}} }\\
    \triangleq & \bb_{a,1}^\intercal\bp\prns{\bh_{t,1}} + \eta_{j,1}^{(a)} +\delta_j.
\end{align*}
Following similar arguments as in the proof of \cref{prop: ols},
when the event $\lambda_{\min}\prns{\hat{\Sigma}_{a,1}(t)}\ge \phi_1 t/4$ holds,
\begin{align*}
    \norm{\hat{\bb}_{a,1}(t) - \bb_{a,1}}\le & \frac{4}{ \phi_1 t}\norm{\sum_{j\in \mathcal{S}_{a,1}(t)} \bp_1(\bh_{j,1})\eta_{j,1}^{(a)} } + \frac{4 x_{\max}}{\phi_1 t}\sum_{j\in \mathcal{S}_{a,1}(t)}\abs{ \delta_j }\\
    \le & \frac{4}{ \phi_1 t}\norm{\sum_{j=1}^t \bp_1(\bh_{j,1})\eta_{j,1}^{(a)} \ind\braces{A_{j,1} = a} } + \frac{4 x^2_{\max} }{\phi_1} \max_{a_2\in[2]} \norm{\hat{\bb}_{a_2,2}(t) - \bb_{a_2, 2}}.
\end{align*}
Therefore,
\begin{align*}
     & \pr\prns{\norm{\hat{\bb}_{a,1}(t) - \bb_{a,1}}> \frac{\chi}{x_{\max}}, \mathcal{M}_t} \\
    \le  & \pr\prns{\norm{\sum_{j=1}^t \bp_1(\bh_{j,1})\eta_{j,1}^{(a)} \ind\braces{A_{j,1} = a} } > \frac{\phi_1 t \chi }{8x_{\max}}} \\
    & + \sum_{a_2\in[2]}\pr\prns{  \norm{\hat{\bb}_{a_2,2}(t) - \bb_{a_2, 2}}>\frac{\phi_1\chi}{8 x_{\max}^3} ,\lambda_{\min}\prns{\hat{\Sigma}_{a_2,2}(t)}\ge \frac{\phi_2 t}{4}}\\
    \le & 2d\exp\prns{-\frac{\phi_1^2 \chi^2 t}{128 d x_\infty^2 x_{\max}^2 \sigma^2}}  + 4d \exp\prns{-\frac{\phi_1^2 \phi_2^2 \chi^2 t}{2048 d x_\infty^2 x_{\max}^6 \sigma^2}}, 
\end{align*}
where the last inequality follows from \citet[theorem 2.19]{wainwright2019high} and \cref{eq: useful}.
\end{myproof}
}

\subsubsection{Proof of Regret Upper Bound.}
\label{section: regret upper bound proofs}
\begin{myproof}[Proof of \cref{thm: regret upper bound}]
Denote our algorithm as $\mathcal{A}$. We have
\begin{align*}
    R_T(\mathcal{A}) = &\sum_{t=1}^T \expect\bracks{\abs{Q_1(\bh_{t,1},1)-Q_1(\bh_{t,1},2)}\ind\braces{A_{t,1}\neq A_{t,1}^*}} + \\
    & \sum_{t=1}^T\expect\bracks{\abs{Q_2(\bh_{t,2}, 1) - Q_2(\bh_{t,2}, 2)}\ind\braces{A_{t,2}\neq \arg\max_{a_2} Q_2(\bh_{t,2},a_2)}} \\
    \le & \underbrace{\sum_{t\in \mathcal{T}_0} 4b_{\max}x_{\max}}_{R^{(1)}} + \underbrace{\sum_{t\not \in \mathcal{T}_0}  \expect\bracks{4b_{\max}x_{\max}\ind\braces{\mathcal{G}_{t-1}^C \cup \mathcal{M}_{t-1}^C }}}_{R^{(2)}}\\
    & + \underbrace{\sum_{t\not \in \mathcal{T}_0}  \expect\bracks{\abs{Q_1(\bh_{t,1},1)-Q_1(\bh_{t,1},2)} \ind\braces{A_{t,1}\neq A_{t,1}^*, \mathcal{G}_{t-1}\cap \mathcal{M}_{t-1} }}}_{R^{(3)}}\\
    & + \underbrace{\sum_{t\not \in \mathcal{T}_0}  \expect\bracks{\abs{Q_2(\bh_{t,2},1)-Q_2(\bh_{t,2},2)}\ind\braces{A_{t,2}\neq \arg\max_{a_2} Q_2(\bh_{t,2},a_2), \mathcal{G}_{t-1} \cap \mathcal{M}_{t-1} }}}_{R^{(4)}}.
\end{align*}
We now control each term of $R^{(1)}$ to $R^{(4)}$.

\paragraph{Step I. Controlling $R^{(1)}$.}
By \cref{lemma: forced pull numbers} and the construction of our forced sampling schedule, {$|\mathcal{T}_0| \le 6 q\log T + 256 q^2$}, so
$$R^{(1)} \le 8 q b_{\max} x_{\max} (3\log T + 128q).$$

\paragraph{Step II. Controlling $R^{(2)}$.}
By \cref{prop: convergence of Tilde Q,lemma: prob on mt c}, when $t\ge 256 q^2$ and $t\in \prns{\cap_{a_1, a_2\in[2]}\mathcal{T}_{(a_1, a_2)}^C} \cup \{2^j q\}_{j\ge 5} $,
\begin{align*}
    \mathbb{P}\prns{\mathcal{G}_{t}^C\cup \mathcal{M}_{t}^C } 
    \le & \mathbb{P}\prns{\mathcal{G}_{t}^C} +\pr\prns{\mathcal{M}_{t}^C}  \\
    \le & \frac{294}{t^2}  + \sum_{m\in[2]} 2d \exp\prns{-\frac{\phi_m t}{48 x_{\max}^2}} \\
    \triangleq & f_0(t).
\end{align*}
Therefore,
\begin{align*}
R^{(2)} = & 4 b_{\max} x_{\max} \sum_{t\not \in \mathcal{T}_0}  \mathbb{P}\prns{\mathcal{G}_{t-1}^C\cup \mathcal{M}_{t-1}^C}  \\
\le & 4 b_{\max} x_{\max} \sum_{t= 256q^2 }^{T} f_0(t-1)\\
\le & 4 b_{\max} x_{\max} \int_{t= 1}^{\infty} f_0(t) dt\\
\le & 24 b_{\max} x_{\max}\prns{49+ \frac{16dx_{\max}^2}{\phi_1} + \frac{16d x_{\max}^2}{\phi_2}}.
\end{align*}

\paragraph{Step III. Controlling $R^{(3)}$.}
For any $t\not\in \mathcal{T}_0$, when $\mathcal{G}_{t-1}$ holds, $\Tilde{Q}_{t-1,1}(\bh_{t,1},a) - \Tilde{Q}_{t-1,1}(\bh_{t,1},3-a)>\Delta_1/2$ implies $Q_1(\bh_{t,1},a)-Q_1(\bh_{t,1},3-a)>0$, so we never make mistakes when we use only the $\Tilde{Q}$ estimators to make decision. Define
\begin{align*}
  D_1(t) = & \braces{\hat{Q}_{t-1,1}(\bh_{t,1},1) - \hat{Q}_{t-1,1}(\bh_{t,1},2)\le 0, Q_1(\bh_{t,1},1)-Q_1(\bh_{t,1},2) \in (0,\Delta_1],  \mathcal{M}_{t-1} } \\
  \supseteq &\Big\{ \abs{\Tilde{Q}_{t-1,1}(\bh_{t,1},1) - \Tilde{Q}_{t-1,1}(\bh_{t,1},2)}\le \frac{\Delta_1}{2}, \hat{Q}_{t-1,1}(\bh_{t,1},1) - \hat{Q}_{t-1,1}(\bh_{t,1},2)\le 0,\\
  & Q_1(\bh_{t,1},1)-Q_1(\bh_{t,1},2)>0, \mathcal{G}_{t-1} \cap \mathcal{M}_{t-1} \Big\},
\end{align*}
$$D_2(t) = \braces{\hat{Q}_{t-1,1}(\bh_{t,1},2) - \hat{Q}_{t-1,1}(\bh_{t,1},1)\le 0, Q_1(\bh_{t,1},2)-Q_1(\bh_{t,1},1) \in (0,\Delta_1],  \mathcal{M}_{t-1} }.$$
We have
$$R^{(3)} \le \sum_{t\not \in \mathcal{T}_0}\expect \bracks{\abs{Q_1(\bh_{t,1},1)-Q_1(\bh_{t,1},2)}\ind\braces{D_1(t) \cup D_2(t)}}.$$
For any $t\not\in \mathcal{T}_0$ and $r = 0,1,2,\dots$, define
$$\mathcal{B}_{1,r}(t) = \braces{2 r \delta_t \le Q_1(\bh_{t,1},1)-Q_1(\bh_{t,1},2) \le 2 (r+1) \delta_t},$$
where $\delta_t$ is a parameter that we will choose later to minimize regret. Since
$$D_1(t) \subseteq \cup_{r=0}^{\lfloor \Delta_1/(2\delta_t) \rfloor} \braces{\hat{Q}_{t-1,1}(\bh_{t,1},1) - \hat{Q}_{t-1,1}(\bh_{t,1},2)\le 0, \mathcal{B}_{1,r}(t),  \mathcal{M}_{t-1}  },$$
we have
\begin{align*}
    &\expect\bracks{(Q_1(\bh_{t,1},1)-Q_1(\bh_{t,1},2)\ind\{D_1(t)\}}\\
    \le & 2\delta_t\sum_{r=0}^{\lfloor \Delta_1/(2\delta_t) \rfloor} (r+1) \mathbb{P}(\hat{Q}_{t-1,1}(\bh_{t,1},1) -\hat{Q}_{t-1,1}(\bh_{t,1},2)\le 0,  \mathcal{M}_{t-1} , \mathcal{B}_{1,r}(t)).
\end{align*}
Moreover, when $\mathcal{B}_{1,r}(t)$ holds, $\hat{Q}_{t-1,1}(\bh_{t,1},1) -\hat{Q}_{t-1,1}(\bh_{t,1},2)\le 0$ implies either $|Q_1(\bh_{t,1},1) - \hat{Q}_{t-1,1}(\bh_{t,1},1)|\ge r\delta_t$ or $|Q_1(\bh_{t,1},2) - \hat{Q}_{t-1,1}(\bh_{t,1},2)|\ge r\delta_t$. By a union bound,
\begin{align*}
 & \mathbb{P} \prns{\hat{Q}_{t-1,1}(\bh_{t,1},1) -\hat{Q}_{t-1,1}(\bh_{t,1},2)\le 0,  \mathcal{M}_{t-1}, \mathcal{B}_{1,r}(t)}   \\
 \le & \sum_{a\in[2]} \mathbb{P} \prns{\abs{Q_1(\bh_{t,1},a) - \hat{Q}_{t-1,1}(\bh_{t,1},a)}\ge r\delta_t,  \mathcal{M}_{t-1} , \mathcal{B}_{1,r}(t)}\\
 \le & \sum_{a\in[2]} \mathbb{P} \prns{\sup_{\mathbf{h}\in \mathcal{H}_1}\abs{Q_1(\mathbf{h},a) - \hat{Q}_{t-1,1}(\mathbf{h},a)}\ge r\delta_t,  \mathcal{M}_{t-1} , \mathcal{B}_{1,r}(t)}\\
 \le & \sum_{a\in[2]} \mathbb{P}\prns{ \sup_{\mathbf{h}\in \mathcal{H}_1}\abs{Q_1(\mathbf{h},a) - \hat{Q}_{t-1,1}(\mathbf{h},a)}\ge r\delta_t,  \mathcal{M}_{t-1} } \mathbb{P}\prns{ \mathcal{B}_{1,r}(t)},
\end{align*}
where the last equality follows from the fact that $\bh_{t,1}$ is independent of  $\max_{\mathbf{h}\in \mathcal{H}_1} |\hat{Q}_{t-1,1}(\mathbf{h},a)-Q_1(\mathbf{h},a)|$ and  $\mathcal{M}_{t-1}$ (depending only on events up to time $t-1$) .

By \cref{prop: convergence of hat Q 1} and \cref{assumption: margin},
\begin{align*}
& \mathbb{P}\prns{ \sup_{\mathbf{h}\in \mathcal{H}_1}\abs{Q_1(\mathbf{h},a) - \hat{Q}_{t-1,1}(\mathbf{h},a)}\ge r\delta_t,  \mathcal{M}_{t-1} } \mathbb{P}\prns{ \mathcal{B}_{1,r}(t)}  \\
\le & \min\braces{1, 6d\exp\prns{-C_1 r^2 \delta_t^2(t-1)} }\mathbb{P}( \mathcal{B}_{1,r}(t)) \\
\le & \gamma_1 (2(r+1)\delta_t)^{\alpha_1}\min\braces{1, 6d\exp\prns{-C_1 r^2 \delta_t^2(t-1)} } .
\end{align*}
If we set $\delta_t = \frac{1}{\sqrt{C_1(t-1)}}$ , we have
\begin{align*}
    &\expect\bracks{\abs{Q_1(\bh_{t,1},1)-Q_1(\bh_{t,1},2)}\ind\{D_1(t) \cup D_2(t)\}}\\
    \le & 8\gamma_1\delta_t\sum_{r=0}^{\lfloor \Delta_1/(2\delta_t) \rfloor} (r+1)  (2(r+1)\delta_t)^{\alpha_1}\min\braces{1, 6d\exp\prns{-C_1 r^2 \delta_t^2(t-1)} }\\
    \le & \underbrace{2^{\alpha_1+3}\gamma_1 C_1^{-\frac{\alpha_1+1}{2}} (t-1)^{-\frac{\alpha_1+1}{2}}\sum_{r=0}^{M} (r+1)^{\alpha_1+1}}_{J_1}  \\
    & + \underbrace{2^{\alpha_1+6} d \gamma_1 C_1^{-\frac{\alpha_1+1}{2}} (t-1)^{-\frac{\alpha_1+1}{2}} \sum_{r=M+1}^{\infty} (r+1)^{\alpha_1+1}\exp\prns{-r^2}}_{J_2}.
\end{align*}
We now control each term in the sum. Note that
$$\sum_{r=0}^{M} (r+1)^{\alpha_1+1} = \sum_{r=1}^{M+1} r^{\alpha_1+1} \le \int_{1}^{M+2} r^{\alpha_1+1} dr = \frac{(M+2)^{\alpha_1+2}-1}{\alpha_1+2} \le \frac{1}{2}(M+2)^{\alpha_1+2},$$
so
$$J_1 \le 2^{\alpha_1+2}\gamma_1 (M+2)^{\alpha_1+2} C_1^{-\frac{\alpha_1+1}{2}}  (t-1)^{-\frac{\alpha_1+1}{2}}.$$
Moreover, when $r\ge M+1 \ge \sqrt{(\alpha_1+4)\log ((\alpha_1+4) d)}$, we have
\begin{align*}
    & r^2 - (\alpha_1+3)\log r - \log d\\
    \ge & (\alpha_1+4)\log ((\alpha_1+4) d) - \frac{1}{2}(\alpha_1+3)(\log (\alpha_1+4)+ \log\log ((\alpha_1+4) d)) - \log d\\
    \ge & (\alpha_1+4)\log ((\alpha_1+4) d) - \frac{1}{2}(\alpha_1+3)(\log ((\alpha_1+4) d)+ \log ((\alpha_1+4) d)) - \log ((\alpha_1+4) d)\\
    = & 0,
\end{align*}
so $(r+1)^{\alpha_1+1} e^{-r^2} \le (2r)^{\alpha_1+1} e^{-r^2} \le 2^{\alpha_1+1}/\prns{d r^2}$. Therefore,
\begin{align*}
    J_2 \le 2^{\alpha_1+6} d \gamma_1 C_1^{-\frac{\alpha_1+1}{2}} (t-1)^{-\frac{\alpha_1+1}{2}} \sum_{r=M+1}^{\infty} \frac{2^{\alpha_1+1}}{d r^2}\le 2^{2\alpha_1+8} \gamma_1 C_1^{-\frac{\alpha_1+1}{2}} (t-1)^{-\frac{\alpha_1+1}{2}}.
\end{align*}
Summing up over $t\not\in \mathcal{T}_0$, we have
\begin{align*}
  R^{(3)} \le  & \sum_{t=256 q^2}^{T} 2^{\alpha_1+2}\gamma_1  C_1^{-\frac{\alpha_1+1}{2}} \prns{(M+2)^{\alpha_1+2}+ 2^{\alpha_1+6}} (t-1)^{-\frac{\alpha_1+1}{2}} \\
  \le &  2^{\alpha_1+2}\gamma_1  C_1^{-\frac{\alpha_1+1}{2}} \prns{(M+2)^{\alpha_1+2}+ 2^{\alpha_1+6}}\int_{t=1}^{T} t^{-\frac{\alpha_1+1}{2}} d t.
\end{align*}
Lastly, by basic calculus we get
$$\int_{t=1}^{T} t^{-\frac{\alpha_1+1}{2}} d t \le
\begin{cases}
\frac{2}{1-\alpha_1}T^{\frac{1-\alpha_1}{2}},& \alpha_1 <1\\
\log T,& \alpha_1 =1 \\
\frac{2}{\alpha_1 - 1},& \alpha_1 >1.
\end{cases}$$

\paragraph{Step IV. Controlling $R^{(4)}$.}
The proof of this part is almost analogous to the proof in Step III except for changing some constants. The only major difference is that, if we let
$$\mathcal{B}_{2,r}(t) = \braces{{2 r \delta_t^{(2)} \le Q_2(\bh_{t,2},1)-Q_2(\bh_{t,2},2) \le 2 (r+1) \delta_t^{(2)}}},$$
since the distribution of $\bh_{t,2}$ depends on which arm we pull in the first stage, we have
\begin{align*}
  \pr(\mathcal{B}_{2,r}(t))
  \le &\sum_{a\in[2]}\mathbb{P} \prns{0<\abs{Q_2(\bh_{t,2}(a),1) - Q_2(\bh_{t,2}(a),2)}\le 2 (r+1) \delta_t^{(2)}}\\
  \le & 2 \gamma_2 \prns{2 (r+1) \delta_t^{(2)}}^{\alpha_2}.
\end{align*}
Following similar arguments as in Step III, we get
$$R^{(4)} \le
\begin{cases}
2^{\alpha_2+4}(1-\alpha_2 )^{-1}\gamma_2  C_2^{-\frac{\alpha_2+1}{2}} ((M+2)^{\alpha_2+2}+ 2^{\alpha_2+4})T^{\frac{1-\alpha_2}{2}},& \alpha_2 <1\\
2^{\alpha_2+3}\gamma_2  C_2^{-\frac{\alpha_2+1}{2}} \prns{(M+2)^{\alpha_2+2}+ 2^{\alpha_2+4}}\log T,& \alpha_2 =1 \\
2^{\alpha_2+4}(\alpha_2 - 1)^{-1}\gamma_2  C_2^{-\frac{\alpha_2+1}{2}} \prns{(M+2)^{\alpha_2+2}+ 2^{\alpha_2+4}},& \alpha_2 >1.
\end{cases}$$

\end{myproof}

\section{DTR Bandit with Delayed Feedback}\label{section: delay}

Because the second stage of a DTR might only occur some time after the first, in many real-life settings we do not get to observe all second-stage outcomes of previous arrivals before a new unit arrives, unlike what is modeled in \cref{section: problem setup}.
In this section, we show that for certain arrival processes, a modified version of \cref{alg} can enjoy similar regret guarantees even when observation of some outcomes are delayed, incurring only an additive penalty due to delays.

We first define sub-exponential tails, which are used to characterize the arrival process.
\begin{definition}[Sub-exponential Random Variables]
  A random variable $X$ with mean $\mu = \expect[X]$ is sub-exponential if there are positive parameters $(\nu, b)$ such that
  $$\expect\bracks{e^{\lambda(X-\mu)}}\le e^{\frac{\nu^2\lambda^2}{2}} \quad \forall |\lambda|<\frac{1}{b}.$$
\end{definition}

Assume that first stage observations arrive in intervals $I_t^1$, which are i.i.d. non-negative sub-exponential random variables with mean $\mu_1$ and parameters $(\nu_1, b_1)$.
That is, the first stage of the $t$-th unit arrives at time $\sum_{t'\leq t}I_{t'}^1$.
A special case is a Poisson arrival process, where the interarrival times are i.i.d. exponential.
Moreover, assume for the $t$-th unit we observe the second stage context with a delay of $I_t^2$ after its first-stage arrival, where $I_t^2$ are i.i.d. non-negative sub-exponential random variables with mean $\mu_2$ and parameters $(\nu_2, b_2)$.
That is, the second stage of the $t$-th unit arrives at time $I_t^2+\sum_{t'\leq t}I_{t'}^1$.
Without loss of generality, assume we pull an arm and observe the outcome immediately after each context arrival -- any delay in this can simply be added into the delay of observing the context.

We now describe our modified algorithm, which we summarize in \cref{alg_delayed}.
{For any timestep $t \not\in \cup_{a_1, a_2 \in[2]} \mathcal{T}_{(a_1, a_2)}$, let $e_t = \max\braces{j<t: j\in \cup_{a_1, a_2 \in[2]} \mathcal{T}_{(a_1, a_2)}} \ge t/2$ be the last timestep before $t$ where we force-pull.}
Similar to \cref{alg}, our proposed algorithm for DTR bandit with delayed feedback maintains two sets of $Q$ estimators, $\Tilde{Q}^{\text{ob}}_{t,m}(\mathbf{h},a)$ and $\hat{Q}^{\text{ob}}_{t,m}(\mathbf{h},a)$, one based only on force-pull samples and the other based on all samples.
There are two major differences from \cref{alg}.
First, the corresponding $\bb$-estimators are based only the samples that have been \textit{observed} by the time index $t$ (which is emphasized by the superscript ``ob'').
Second, when we make decisions at time $t$, our tilde estimator $\Tilde{Q}^{\text{ob}}_{e_t,m}(x,a)$ is based on samples observed before $e_t$ instead of $t-1$. This second change is in order to be able to guarantee a similar result as \cref{lemma: prob on mt c}.

\begin{algorithm}[t!]%
    \caption{\textsc{DelayedDTRBandit}}
    \label{alg_delayed}
    \begin{algorithmic}[1]
    \item \textbf{Input parameters:} $\Delta_1, \Delta_2, q$.
    \FOR{arrival $(t,m) \in \mathbb{N} \times [2]$}
                \IF{$t \in \mathcal{T}_{(a_1, a_2)}$ for $a_1, a_2 \in [2]$}
                \STATE{pull $A_{t,m} = a_m$}
                \ELSE
                \STATE{compute $\hat{\bb}^{\text{ob}}_{a,m}(t-1)$, $\Tilde{\bb}^{\text{ob}}_{a,m}(e_t)$ for $a, m \in [2]$}
                \IF{$\abs{\Tilde{Q}^{\text{ob}}_{e_t,m}(\bh_{t,m},1)-\Tilde{Q}^{\text{ob}}_{e_t,m}(\bh_{t,m},2)}>\Delta_m/2$}
                \STATE{pull $A_{t,m} = \arg\max_{a=1,2} \Tilde{Q}^{\text{ob}}_{e_t,m}(\bh_{t,m},a)$}
                \ELSE
                \STATE{pull $A_{t,m} = \arg\max_{a=1,2} \hat{Q}^{\text{ob}}_{t-1,m}(\bh_{t,m},a)$}
                \ENDIF
                \ENDIF
    \ENDFOR
    \end{algorithmic}
\end{algorithm}

\cref{alg_delayed} enjoys the same order of regret in terms of $T$ and $d$ dependence as in \cref{thm: regret upper bound}, as is shown below in \cref{thm: delay}. In particular, the additional regret due to delays is additive.
\begin{theorem}\label{thm: delay}
  Let $\hat{\mathcal A}'$ denote our algorithm described in \cref{alg_delayed}. Suppose \cref{assumption: bounded,assumption: margin,assumption: diversity} hold. Then there exists a constant $\Tilde C$ such that, if $q\ge\Tilde C$, then the expected regret of our algorithm is bounded as follows:
  \begin{equation}\label{eq:upperboundrate_delay}R_T(\hat{\mathcal{A}}') =
  \begin{cases}
  O\prns{d^{\frac{\alpha+1}{2}}(\log d)^{\frac{\alpha+2}{2}} T^{\frac{1-\alpha}{2}} +  d \log d \log T + (d\log d)^2 + \max\braces{\frac{\nu_1^4}{\mu_1^4 }, \frac{b_1^2}{\mu_1^2 }, \frac{\nu_2^4}{\mu_1^4}, \frac{b_2^2}{\mu_1^2 }}+\frac{\mu_2}{\mu_1}} ,& \alpha <1\\
  O \prns{d (\log d)^{\frac{3}{2}} \log T + (d\log d)^2 + \max\braces{\frac{\nu_1^4}{\mu_1^4 }, \frac{b_1^2}{\mu_1^2 }, \frac{\nu_2^4}{\mu_1^4}, \frac{b_2^2}{\mu_1^2 }}+\frac{\mu_2}{\mu_1}},& \alpha =1 \\
  O\prns{d \log d \log T + (d\log d)^2 +  d^{\frac{\alpha+1}{2}}(\log d)^{\frac{\alpha+2}{2}}+ \max\braces{\frac{\nu_1^4}{\mu_1^4 }, \frac{b_1^2}{\mu_1^2 }, \frac{\nu_2^4}{\mu_1^4}, \frac{b_2^2}{\mu_1^2 }}+\frac{\mu_2}{\mu_1}},& \alpha >1.
  \end{cases}\end{equation}
  where the $O$ notation hides anything that is constant in $T,d, \mu_1,\mu_2, \nu_1, \nu_2, b_1, b_2$.
  \end{theorem}
\begin{proof}[Proof Sketch of \cref{thm: delay}]
Define the event
\begin{align*}
  \mathcal O_t = \braces{\bh_{1,2}, \dots, \bh_{\lfloor t/2\rfloor,2} \text{ arrive before } \bh_{t,1}}.
\end{align*}
We have
\begin{align*}
  \mathbb{P}\prns{\mathcal O_t^C} & \le \sum_{i=1}^{\lfloor t/2 \rfloor} \mathbb{P}\prns{\sum_{j=i+1}^t I^1_j < I^2_i} \\
  & \le \frac{t}{2} \mathbb{P}\prns{\sum_{j=\lfloor t/2 \rfloor+1}^t I^1_j < I^2_1} \\
  & \le \frac{t}{2} \mathbb{P}\prns{\sum_{j=\lfloor t/2 \rfloor+1}^t I^1_j <\frac{1}{2}\prns{t-\lfloor t/2 \rfloor}\mu_1} + \frac{t}{2} \mathbb{P}\prns{I^2_1>\frac{t-1}{4}\mu_1}.
\end{align*}
By \citet[equation (2.18)]{wainwright2019high}, we know that, when $t\ge 8\mu_2/\mu_1+2$,
\begin{align*}
    \mathbb{P}\prns{\sum_{j=\lfloor t/2 \rfloor+1}^t I^1_j <\frac{1}{2}(t-\lfloor t/2 \rfloor)\mu_1}\le \exp\prns{-\min\braces{\frac{\mu_1^2 t}{16\nu_1^2}, \frac{\mu_1 t}{8b_1}}},
\end{align*}
\begin{align*}
  \mathbb{P}\prns{I^2_1>\frac{t-1}{4}\mu_1} & \le  \exp\prns{-\min\{\frac{((t-1)\mu_1/4-\mu_2)^2}{2\nu_2^2}, \frac{(t-1)\mu_1/4-\mu_2}{2b_2}\}}
  \\ & \le \exp\prns{-\min\braces{\frac{\mu_1^2t^2}{128\nu_2^2}, \frac{\mu_1t}{16b_2}}},
\end{align*}
which implies that for $t\ge 8\mu_2/\mu_1+2$,
\begin{align*}
  \mathbb{P}\prns{\mathcal O_t^C} \le t\exp\prns{-\min\braces{\frac{\mu_1^2 t}{16\nu_1^2}, \frac{\mu_1 t}{8b_1}, \frac{\mu_1^2t^2}{128\nu_2^2}, \frac{\mu_1 t}{16b_2}}}.
\end{align*}
This means that by the time the t-th first-stage context arrives, with high probability the decision maker would have observed outcomes for both stages of the first $\lfloor t/2 \rfloor$ arrivals, which is intuitively sufficient to get same-order estimation accuracy for the $Q$ estimators as in the situation where we have prompt observations.
At time $t$, we can use similar arguments as in the proof of \cref{thm: regret upper bound} to show that
\begin{align*}
  & \mathbb P (\mathcal G_{e_t}^C) \le O\prns{t^{-2}} + O\prns{te^{-c\min\braces{\frac{\mu_1^2 }{\nu_1^2}, \frac{\mu_1 }{b_1}, \frac{\mu_1^2}{\nu_2^2}, \frac{\mu_1 }{b_2}}\prns{t\wedge t^2}}},\\
  & \mathbb P (\mathcal M_{t-1}^C) \le O\prns{t^{-2}} + O\prns{de^{-c't}} + O\prns{te^{-c'' \min\braces{\frac{\mu_1^2 }{\nu_1^2}, \frac{\mu_1 }{b_1}, \frac{\mu_1^2}{\nu_2^2}, \frac{\mu_1 }{b_2}}\prns{t\wedge t^2}}}
\end{align*}
for some constants $c, c', c''$ independent of $t,d, \mu_1,\mu_2, \nu_1, \nu_2, b_1, b_2$,
where the extra term $O\prns{t\exp\prns{-c \min\braces{\mu_1^2/\nu_1^2, \mu_1/b_1,\mu_1^2/\nu_2^2, \mu_1/b_2}\prns{t\wedge t^2}}}$ accounts for situations where event $\mathcal O_{e_t}$ fails.
The rest of the proof is similar to the proof of \cref{thm: regret upper bound}.
\end{proof}

\section{Extension to Multiple Arms and to Allowing Suboptimal Arms}
\label{section: k arms}
In this section, we briefly discuss how our algorithm and analysis can be modified to solve the multiple-armed DTR bandits and also allowing for the possibility of suboptimal arms. We index our arms in the first stage as $1,\dots,K_1$ and in the second stage as $1,\dots,K_2$.
We define ${Q}_{t,m}$ as in the $2$-armed case except replacing $\max_{a\in[2]}$ with $\max_{a\in[K_m]}$.
As before, we assume that both $Q$-functions are linear, and the parameters are well bounded as in \cref{assumption: bounded}.
Moreover, we make slight modification to \cref{assumption: margin,assumption: diversity} as follows:
\begin{assumption}[Margin Condition]\label{assumption: margin k arms}
There exist positive constants $\gamma_1, \gamma_2, \alpha_1, \alpha_2$ such that for any $\kappa>0$,
\begin{align*} %
   &\mathbb{P}\prns{0<\abs{Q_1(\bx_{t,1},i)-Q_1(\bx_{t,1},j)}\le \kappa}\le \gamma_1 \kappa^{\alpha_1} \quad\forall i,j\in [K_1],
\\
    &\mathbb{P}\prns{0<\abs{Q_2(\bh_{t,2}(a),i) - Q_2(\bh_{t,2}(a),j)}\le \kappa}\le \gamma_2 \kappa^{\alpha_2}\quad\forall a\in[K_1],i,j\in [K_2].
\end{align*}
\end{assumption}

\begin{assumption}[Positive-definiteness of Design Matrices, Allowing for Suboptimal Arms]\label{assumption: diversity k arms (with suboptimal arms)}
Let $\braces{\mathcal{K}_{sub,1}, \mathcal{K}_{opt,1}}$  and $\braces{\mathcal{K}_{sub,2}, \mathcal{K}_{opt,2}}$ be partitions of $[K_1]$ and $[K_2]$, respectively.
There exist positive constants $\Delta_1, \Delta_2$ that satisfy the following properties:
\begin{itemize}
  \item Sub-optimal arms satisfy $Q_m(\mathbf{h},a) < \max_{i\in[K_m]}Q_m(\mathbf{h},i) - \Delta_m, \quad \forall m\in[2], \mathbf{h}\in \mathcal{H}_m, a\in \mathcal{K}_{sub,m}.$
  \item Let $U_{a,m} \triangleq \braces{\mathbf{h}\in \mathcal{H}_m: Q_m(\mathbf{h},a) > \max_{i\neq a}Q_m(\mathbf{h},i)+\Delta_m}$, $\forall m \in [2], a\in \mathcal{K}_{opt,m}$. There exist $\phi_1, \phi_2 >0$ such that
  \begin{enumerate}
    \item $\lambda_{\min}\prns{\expect[\bp_1(\bx_{t,1})\bp_1^\intercal(\bx_{t,1})\ind\{\bx_{t,1}\in U_{a,1}\}]}\ge \phi_1 , \forall a\in \mathcal{K}_{opt,1}$ \label{assumption3-2: 2}
    \item $\lambda_{\min}\prns{\sum_{i\in\mathcal{K}_{opt,1}}\expect\bracks{\bp_2(\bh_{t,2}(i))\bp_2^\intercal(\bh_{t,2}(i)) \ind\{\bh_{t,2}(i) \in U_{a,2}, \bx_{t,1}\in U_{i,1}\}}}\ge \phi_2, \forall a\in \mathcal{K}_{opt,2}$. \label{assumption3-2: 4}
  \end{enumerate}
\end{itemize}
\end{assumption}
It's easy to see that \cref{assumption: diversity k arms (with suboptimal arms)} naturally imply that there exist $p_1, p_2>0$ such that
$$\mathbb{P}\prns{\bx_{t,1}\in U_{a,1}} \ge p_1,\quad \forall a\in \mathcal{K}_{opt,1},$$ $$\sum_{i\in\mathcal{K}_{opt,1}}\mathbb{P}\prns{\bh_{t,2}(i) \in U_{a,2}, \bx_{t,1}\in U_{i,1}}\ge p_2, \quad \forall a\in \mathcal{K}_{opt,2}.$$
In contrast, this is not required for the remaining arms in $\mathcal{K}_{sub,1},\mathcal{K}_{sub,2}$, which are allowed to be suboptimal everywhere as long as they are suboptimal by a margin.

We now proceeds to describe our algorithm. Our forced pull schedule is now 
{
for $a_1\in[K_1], a_2\in[K_2]$,
\begin{equation}\label{eq: forced samples K}
   \mathcal{T}_{(a_1, a_2)} \triangleq \braces{(2^n-1)K_1 K_2q+j~:~j=q(K_2 a_1 + a_2 - K_2 -1)+1,\dots, q (K_2 a_1 + a_2 - K_2),\ n=0,1,\dots}.
\end{equation}
}
Both $\Tilde{Q}_{t,m}$ and $\hat{Q}_{t,m}$ can be estimated in the same way as in the $2$-armed case except replacing $\max_{a\in[2]}$ with $\max_{a\in[K_m]}$. In each stage, we first filter out seemingly suboptimal arms based on $\Tilde{Q}_{t,m}$, then pull the arm with highest $\hat{Q}_{t,m}$ estimate among the remaining arms. The detailed algorithm is summerized in \cref{alg: k arms}.

\begin{algorithm}[t!]%
    \caption{\textsc{K-armed DTRBandit}}
    \label{alg: k arms}
    \begin{algorithmic}[1]
    \item \textbf{Input parameters:} $\Delta_1, \Delta_2, q$.
    \FOR{$t = 1,2,\dots$}
                \IF{$t \in \mathcal{T}_{(a_1, a_2)}$ for $a_1\in[K_1], a_2\in[K_2]$}
                \STATE{pull $A_{t,1} = a_1,\,~ A_{t,2} = a_2$}
                \STATE{observe rewards and compute $\hat{\bb}_{a,m}(t)$, $\Tilde{\bb}_{a,m}(t)$,  for $m \in[2]$ and $a\in [K_m]$}
                \ELSE
                \FOR{$m = 1,2$}
                \STATE set $\hat{\mathcal{K}}_{opt,m} =\braces{a\in [K_m]: \max_{i\in[K_m]}\Tilde{Q}_{t-1,m}(\bh_{t,m},i)-\Tilde{Q}_{t-1,m}(\bh_{t,m},a)\le\frac{\Delta_m}{2}}$
                \STATE{pull $A_{t,m} = \arg\max_{a\in \hat{\mathcal{K}}_{opt,m}} \hat{Q}_{t-1,m}(\bh_{t,m},a)$}
                \STATE{observe rewards and compute $\hat{\bb}_{a,m}(t)$ for $a\in [K_m]$}
                \ENDFOR\\
                \ENDIF
    \ENDFOR
    \end{algorithmic}
\end{algorithm}

Finally, we briefly explain how our analysis extends easily to show that \cref{alg: k arms} has low regret.
For $\forall t\in [T]$, define
\begin{align}
  & \mathcal{G}_t = \braces{\max_{m\in[2]}\max_{a\in[K_m]}\sup_{\mathbf{h}\in \mathcal{H}_m} \abs{\Tilde{Q}_{t,m}(\mathbf{h},a)-Q_m(\mathbf{h},a)} \le \frac{\Delta_m}{4} },\\
   & {\mathcal{M}_t = \braces{\max_{m\in[2]}\min_{a\in \mathcal{K}_{opt,m}}\lambda_{\min}\prns{\hat{\Sigma}_{a,m}(t)}\ge \frac{\phi_m t}{4}}}, \\
   &{\mathcal{L}_t = \braces{\max_{a\in[K_2]}\norm{\hat{\bb}_{a,2}(t)- \bb_{a,2}}\le \frac{\Delta_2}{2x_{\max}}}}.
\end{align}
Compared to \cref{eq: event M}, here $\mathcal{M}_t$ only requires the minimum eigenvalues of the design matrices for \textit{optimal arms} to grow like $\Omega(T)$.
Moreover, we define a new event $\mathcal{L}_t$ that requires an additional estimation accuracy guarantee on $\max_{a\in[K_2]}\norm{\hat{\bb}_{a,2}(t)- \bb_{a,2}}$.
Following similar analysis as in \cref{section: proof tilde,section: proof hat}, as long as $q$ is chosen large enough, we have $\mathbb{P}\prns{\mathcal{G}_t^C \cup \mathcal{M}_t^C \cup \mathcal{L}_t^C} = O(1/t^2).$
Moreover, when $\mathcal{G}_t \cap \mathcal{M}_t \cap \mathcal{L}_t$ holds, for $\forall a\in \mathcal{K}_{opt,2}$, $\norm{\hat{\bb}_{a,2}(t)- \bb_{a,2}}$ has a sub-Gaussian tail.

{
Controlling $\norm{\hat{\bb}_{a,1}(t)- \bb_{a,1}}$ is more subtle due to suboptimal arms.
Recall that
\begin{align*}
     \hat{\bb}_{a, 1}(t) & = \hat{\Sigma}_{a,1}^{-1}(t) \prns{\sum_{j\in \mathcal{S}_{a,1}(t)} \bp_1(\bx_{j,1})\hat{Y}_j  } ,
\end{align*}
where
\begin{align*}
    \hat{Y}_j = &\max_{a_2\in[K_2]} \hat{\bb}_{a_2, 2}^\intercal(t) \bp_2\prns{\bh_{j,2}} \\
    = & \bb_{a,1}^\intercal\bp\prns{\bh_{t,1}} + \eta_{j,1}^{(a)} +\bracks{\max_{a_2\in[K_2]} \hat{\bb}_{a_2,2}^\intercal(t)\bp_2\prns{\bh_{j,2}} - \max_{a_2\in[K_2]} \bb_{a_2,2}^\intercal\bp_2\prns{\bh_{j,2}} }.
\end{align*}
Thus, proving convergence of $\hat{\bb}_{a, 1}(t)$ depends on controlling $\max_{a_2\in[K_2]} \hat{\bb}_{a_2,2}^\intercal(t)\bp_2\prns{\bh_{j,2}} - \max_{a_2\in[K_2]} \bb_{a_2,2}^\intercal\bp_2\prns{\bh_{j,2}} $, but we can only guarantee sub-Gaussian convergence of $\hat{\bb}_{a_2,2}(t)$ for optimal arms in stage $2$.
This can be solved by the following arguments.
By \cref{assumption: diversity k arms (with suboptimal arms)}, for all $ a'\in \mathcal{K}_{sub,2},  a\in \mathcal{K}_{opt,1}, j\in \mathcal{S}_{a,1}(t)$,
$$\bb_{a',2}^\intercal \bp_2\prns{\bh_{j,2}(a)} < \max_{a_2\in[K_2]}\bb_{a_2,2}^\intercal \bp_2(\bh_{j,2}(a)) - \Delta_2.$$
Moreover, when $\mathcal{L}_t$ holds, we have
$$\max_{a_2\in [K_2],  a\in \mathcal{K}_{opt,1}, j\in \mathcal{S}_{a,1}(t)}\abs{\bb_{a_2,2}^\intercal \bp_2\prns{\bh_{j,2}(a)} - \hat{\bb}_{a_2,2}^\intercal \bp_2\prns{\bh_{j,2}(a)}} \le \frac{\Delta_2}{2} .$$
These two inequality imply that for any $a'\in \mathcal{K}_{sub,2}$,
\begin{align*}
    \hat{\bb}_{a', 2}^\intercal(t) \bp_2\prns{\bh_{j,2}(a)} \le & \bb_{a', 2}^\intercal(t) \bp_2\prns{\bh_{j,2}(a)} + \frac{\Delta_2}{2} \\
    < & \max_{a_2\in[K_2]}\bb_{a_2,2}^\intercal \bp_2(\bh_{j,2}(a)) - \frac{\Delta_2}{2}\\
    \le & \max_{a_2\in[K_2]} \hat{\bb}_{a_2,2}^\intercal \bp_2\prns{\bh_{j,2}(a)},
\end{align*}
so
\begin{align*}
  \max_{a_2\in[K_2]} \hat{\bb}_{a_2,2}^\intercal \bp_2\prns{\bh_{j,2}(a)} = \max_{a_2\in\mathcal{K}_{opt,2}} \hat{\bb}_{a_2,2}^\intercal \bp_2\prns{\bh_{j,2}(a)} ,
 \end{align*}
and 
\begin{align*}
  & \max_{a_2\in[K_2]} \hat{\bb}_{a_2,2}^\intercal(t)\bp_2\prns{\bh_{j,2}} - \max_{a_2\in[K_2]} \bb_{a_2,2}^\intercal\bp_2\prns{\bh_{j,2}} \\
  = & \max_{a_2\in\mathcal{K}_{opt,2}} \hat{\bb}_{a_2,2}^\intercal \bp_2\prns{\bh_{j,2}(a)} - \max_{a_2\in\mathcal{K}_{opt,2}} \bb_{a_2,2}^\intercal\bp_2\prns{\bh_{j,2}}\\
  \le & \max_{a_2\in\mathcal{K}_{opt,2}} \norm{\hat{\bb}_{a_2,2}(t)- \bb_{a_2,2}} x_{\max}.
 \end{align*}
Thus, when $\mathcal{G}_t \cap \mathcal{M}_t \cap \mathcal{L}_t$ holds, for all $a\in \mathcal{K}_{opt,1}$, we can prove a similar sub-Gaussian convergence for $\norm{\hat{\bb}_{a,1}(t)- \bb_{a,1}}$ as in \cref{prop: convergence of hat Q 1}.}
Besides, the estimation accuracy guaranteed by $\mathcal{G}_t$ ensures that all suboptimal arms are ruled out of $\hat{K}_{opt,m}$ in both stages, so we do not need to care about convergence of $\hat{\bb}_{a,m}(t)$ for suboptimal arms.
The rest of the proof for the regret upper bound then follows similarly as in \cref{section: regret upper bound proofs}, and we omit the details here.

\end{document}